# Short-Term Electricity Demand Forecasting for New England: A Comprehensive Machine Learning Benchmark with Weather, Calendar, and COVID-19 Indicators

**Reza Ghanavati [1,*], Behrooz Mosallaei [2]**

[1]*Department of Chemical and Materials Engineering, New Mexico State University*

[2]*Department of Electrical and Communications Engineering, New Jersey Institute of Technology*

*corresponding author email: rgh@nmsu.edu

## Abstract

Accurate short-term electricity demand forecasting is critical for reliable power system operation, energy market planning, and infrastructure optimization. This paper benchmarks ten machine learning models for daily electricity demand forecasting across the New England ISO (February 2020 - March 2023). The models span four families: tabular gradient-boosted trees (Random Forest, LightGBM, CatBoost, XGBoost), standalone neural architectures (LSTM, Transformer encoder), and hybrid Transformer+tree variants (Hybrid XGBoost, Hybrid LightGBM, Hybrid CatBoost, Hybrid RF). All models use meteorological data from six cities, calendar and holiday effects, autoregressive demand lags, and COVID-19 epidemiological variables. Hyperparameter optimization uses Optuna (300 trials, multivariate TPE, seed=42) under a leakage-free 70/15/15 chronological split. CatBoost achieves the best test performance: RMSE 8316 MWh, MAPE 1.87%, R-squared 0.917, followed by XGBoost (9066 MWh, R-squared 0.901), Hybrid CatBoost (9068 MWh, R-squared 0.901), and Hybrid XGBoost (9208 MWh, R-squared 0.898). Standalone neural architectures perform substantially worse (Transformer: 21294 MWh; LSTM: 22808 MWh), confirming the Transformer's role as a feature extractor rather than an end-to-end forecaster. An ablation on CatBoost shows that demand lags are the dominant predictor: removal degrades RMSE from 8316 to 11310 MWh (+36%), while weather and calendar features alone achieve an R-squared of 0.864. Removing COVID-19 features improves test RMSE by 1.7% while reducing training RMSE by 17.3%, a signature of temporal validity decay. SHAP analysis confirms this: 3 of 8 COVID features rank higher on the post-acute test set than during pandemic-active training, indicating the model over-applies stale pandemic patterns after behavioral adaptation was complete by August 2022.

## 1. Introduction

Electricity is a critical infrastructure resource essential to modern economies, public health systems, and daily life. Unlike most commodities, electricity cannot be stored at scale, which means supply and demand must be balanced in real time across transmission networks. Accurate forecasting of electricity demand is therefore a foundational requirement for power system operators, enabling cost-efficient generation scheduling, transmission congestion management, reserve allocation, and energy market clearing.[1-4]

Electricity demand forecasts are broadly categorized by their time horizon. Long-term forecasting (months to years ahead) informs capacity planning, infrastructure investment, and regulatory decisions.[5-9] Medium-term forecasting (days to weeks) supports fuel procurement, maintenance scheduling, and bilateral contract pricing.[8-13] Short-term forecasting (hours to a few days ahead) is the most operationally critical category, directly driving real-time dispatch, unit commitment, and intraday market operations.[11,13-17] In short-term forecasting, daily horizon models that predict total demand one day ahead play a particularly important role because they govern the day-ahead energy market, which in most regions accounts for the majority of traded electricity volume.[18-20]

A wide variety of methods have been applied to short-term electricity demand forecasting. Classical statistical approaches, including autoregressive integrated moving average (ARIMA) models and exponential smoothing, provided the foundation for early forecasting systems and remain competitive baselines in stable demand environments.[21-23] Machine learning methods, in particular gradient-boosted decision trees, support vector regression, and random forests, substantially improved accuracy by capturing nonlinear relationships between demand and meteorological or calendar drivers without requiring explicit parametric assumptions.[24-26] More recently, deep learning architectures, including long short-term memory (LSTM) networks, convolutional neural networks (CNNs), and Transformer models, have demonstrated strong performance on load forecasting tasks by learning temporal dependencies directly from raw sequential data.[27-34]

Among these approaches, hybrid architectures that combine a deep learning component for temporal feature extraction with a gradient-boosted model for final prediction have attracted growing interest. The motivation is to exploit complementary strengths: deep sequence models such as Transformers capture long-range temporal dependencies and complex interaction patterns that are difficult to engineer manually, while gradient boosting provides robust regularization, strong performance on tabular data, and resistance to overfitting on moderate-sized datasets. Several recent studies have demonstrated that pairing Transformer-derived embeddings with XGBoost or similar gradient-boosted models yields competitive or superior accuracy relative to end-to-end deep learning on load forecasting tasks.[35-38] The present study adopts this hybrid design philosophy as one of ten model families evaluated in a systematic benchmark, using a Transformer encoder as a temporal feature extractor whose embeddings are concatenated with tabular features and passed to gradient-boosted tree models tuned via Bayesian hyperparameter optimization.

The COVID-19 pandemic introduced structural disruption to electricity demand patterns, challenging conventional forecasting models trained on pre-pandemic data. Widespread lockdowns, remote-work mandates, and commercial closures sharply altered the weekday–weekend demand differential and shifted load from commercial and industrial sectors toward residential consumption, producing demand profiles that diverged substantially from historical seasonal norms.[39,40] Several studies have incorporated epidemiological indicators, including case counts, mobility indices, and policy stringency measures, as additional features to capture pandemic-driven behavioral shifts and improve forecast accuracy during the acute phase.[41-44] The quantitative contribution of such variables, however, remains context-dependent and has rarely been examined beyond the initial lockdown period, leaving open the question of whether epidemiological features retain predictive value as behavioral adaptation progresses.

The present study focuses on the New England electricity market, operated by ISO New England (ISO-NE), a regional transmission organization overseeing the six-state region encompassing Connecticut, Maine, Massachusetts, New Hampshire, Rhode Island, and Vermont. New England presents a particularly demanding forecasting environment: the region's demand profile is highly weather-sensitive, with cold winters driving substantial electric heating loads and humid summers generating sharp cooling peaks across a geographically diverse service territory. These meteorological extremes, combined with strong weekly periodicity, public holiday effects, and the structural disruptions of the COVID-19 pandemic spanning 2020–2022, make New England a rich and challenging case study. This paper benchmarks ten machine learning models for short-term daily demand forecasting across the New England ISO control area over the period February 2020 to March 2023, integrating multi-city weather observations from six cities spanning all six states, calendar indicators, autoregressive demand lags, and epidemiological variables derived from COVID-19 case and death records. Hyperparameters are optimized using Optuna [45], a Bayesian optimization framework employing the Tree-structured Parzen Estimator, enabling efficient search over the joint model configuration space. A systematic ablation study and SHapley Additive exPlanations (SHAP) [46] interpretability analysis are conducted to quantify each feature group's contribution and to characterize the temporal validity of COVID-19 signals across the full arc from acute pandemic disruption to post-acute stabilization.

The main contributions of this work are:

- Development of a comprehensive feature engineering pipeline incorporating multi-city weather observations from six cities across New England, rolling-window COVID-19 statistics, calendar and seasonal indicators, and autoregressive demand lags.
- A comprehensive benchmark of ten models, tabular (Random Forest, LightGBM, CatBoost, XGBoost), sequential neural (LSTM, Transformer encoder), and hybrid Transformer+tree variants (Hybrid XGBoost, Hybrid LightGBM, Hybrid CatBoost, Hybrid RF), optimized under identical conditions (300 Optuna trials, seed=42) to identify the best-performing architecture for this forecasting task.
- Rigorous leakage-free training methodology: the StandardScaler is fitted exclusively on training data, and all splits are chronological, preserving temporal ordering.
- Bayesian hyperparameter optimization with a multivariate TPE sampler and MedianPruner over 300 trials (seed=42), achieving systematic convergence across all ten models under identical conditions.
- Interpretability analysis using SHAP to quantify the contribution of each feature group, with particular attention to COVID-19 variables.

## 2. Data Description

### 2.1. Study Region

The study region is the New England ISO control area, encompassing the states of Connecticut, Maine, Massachusetts, New Hampshire, Rhode Island, and Vermont. The analysis covers the period from February 1, 2020, through March 23, 2023, a span of approximately 1,147 days that includes the full arc of the COVID-19 pandemic in the United States, from the initial outbreak through the post-Omicron stabilization period.

### 2.2. Electricity Demand Data

Daily electricity demand data, measured in megawatt-hours (MWh), are obtained from ISO New England's public data portal.[47] The target variable represents the aggregated daily demand for the entire New England control area. Over the study period, daily demand ranges from approximately 225,000 MWh during mild spring conditions to over 500,000 MWh during summer heat waves and cold winter peaks, with a mean of approximately 325,000 MWh and a standard deviation of approximately 60,000 MWh.

### 2.3. Weather Data

Meteorological data are obtained from Weather Underground [48] for six cities distributed across New England: Boston (MA), Bridgeport (CT), Burlington (VT), Concord (NH), Portland (ME), and Providence (RI). This multi-city approach captures the spatial heterogeneity of weather conditions across the region. For each city, 16 daily observations are collected: maximum, average, and minimum values of temperature, dew point, humidity, wind speed, and atmospheric pressure, plus total precipitation. This yields 96 raw weather features (6 cities × 16 variables) before feature engineering.

### 2.4. Calendar and Holiday Effects

Calendar effects are incorporated using binary indicators representing four day types: holidays, weekends, standard workdays, and weekdays falling between a holiday and a weekend ("bridge days"). These indicators are derived from an external custom calendar dataset aligned to the New England region and merged by date. Additionally, calendar features computed from the date itself include the month of the year, day of the week (Monday = 0 through Sunday = 6), day of the year, meteorological season (winter, spring, summer, fall), and binary peak-season flags for summer (June–August) and winter (December–February).

### 2.5. COVID-19 Epidemiological Data

Daily COVID-19 case and death counts aggregated for the New England region are obtained from The New York Times COVID-19 data repository [49], which compiles state-level reports from public health authorities. From the raw case and death series, eight epidemiological features are derived: raw daily cases and deaths, 7-day and 14-day rolling means of cases and deaths (to smooth reporting noise and capture trend momentum), and daily percentage growth rates for both cases and deaths. Infinity values arising from zero-division in growth rate calculations are replaced with NaN and subsequently handled by the standard scaling step.

## 3. Feature Engineering

Feature engineering plays a central role in the proposed framework. Beyond the raw meteorological and epidemiological variables described above, the following additional features are constructed.

Autoregressive demand features are included using lags of 1, 7, 14, and 30 days to capture short-term persistence, weekly seasonality, and monthly patterns. The 1-day lag is the single most predictive feature in the model, as confirmed by SHAP analysis. In hybrid architectures, all 118 standardized features (including the demand lags) are used as input channels to the Transformer encoder.

All features are aligned temporally by merging on a common date index. Rows containing missing values resulting from lag construction (the first 30 days) are removed. The final feature matrix, after merging all data sources and constructing lag and calendar features, contains 118 input features per day.

## 4. Methodology

### 4.1. Overview

The benchmark evaluates ten model families across four categories. Four purely tabular models (Random Forest, LightGBM, CatBoost, XGBoost) operate directly on the 118-feature engineered matrix. Two standalone sequential neural architectures (LSTM, Transformer encoder) process the raw time-series channels without tabular inputs. Four hybrid models concatenate a Transformer encoder's 32-dimensional embedding with the 118 tabular features (150 total inputs) before passing the combined vector to a gradient-boosted regressor (Hybrid XGBoost, Hybrid LightGBM, Hybrid CatBoost, Hybrid RF). All ten models share the same feature engineering pipeline, chronological 70/15/15 train-validation-test split, and 300-trial Optuna search under identical conditions (seed=42), ensuring a fair and reproducible comparison.

### 4.2. Data Preprocessing

All features are standardized using StandardScaler. Critically, the scaler is fitted exclusively on the training set and then applied to the validation and test sets, preventing any leakage of future statistics into the model. The chronological 70/15/15 split yields 770 training days, 165 validation days, and 165 test days. All random operations are seeded with seed 42 (applied to Python’s random module, NumPy, and PyTorch) to ensure full reproducibility.

### 4.3. Transformer Temporal Encoder

The Transformer encoder processes sliding windows of length L = 30 days over all 118 standardized input features, yielding a 118-channel input sequence. Each window is projected to an embedding dimension of d = 32 via a learned linear layer. Sinusoidal positional encoding is added to inject temporal ordering information, which vanilla Transformer encoders otherwise ignore. The encoder consists of two TransformerEncoderLayer blocks with 4 attention heads, a feedforward expansion ratio of 4, and a dropout rate of 0.1. Adaptive average pooling collapses the sequence dimension to produce a fixed 32-dimensional embedding vector per sample.

The encoder is trained end-to-end with a mean squared error loss using the Adam optimizer (learning rate $10^{-3}$), a ReduceLROnPlateau scheduler (patience = 5, factor = 0.5), and an early-stopping rule with patience of 10 epochs. The best encoder weights (minimum validation loss) are restored before generating embeddings. The 32 embedding dimensions are then concatenated with the full tabular feature vector and passed to gradient-boosted regressor.

### 4.4. Gradient Boosting Model

Each gradient-boosted tree model (XGBoost, LightGBM, CatBoost, and their hybrid variants) uses a squared error regression objective. XGBoost is configured with shallow trees (max_depth searched over 3–7) and L1/L2 regularization. CatBoost uses symmetric trees with depth searched over 2–8. LightGBM uses gradient-based one-side sampling. All three apply early stopping with a patience of 50 rounds on validation RMSE, preventing overfitting by halting training when validation performance ceases to improve. Random Forest models use the same feature set without early stopping.

Hyperparameters for all ten models are tuned using Optuna with a multivariate Tree-structured Parzen Estimator (TPE) sampler over 300 trials (seed=42). The multivariate sampler explicitly models the joint distribution of hyperparameters, capturing interactions between parameters such as learning rate and number of estimators. A MedianPruner discards clearly unpromising trials early, focusing computational budget on promising regions of the search space. Table 1 reports the optimal hyperparameters for the two best-performing tabular models (CatBoost and XGBoost). Full hyperparameter tables for XGBoost, Hybrid Transformer+CatBoost, and Hybrid Transformer+XGBoost are provided in the Electronic Supplementary Information (Tables S1–S3).

**Table 1:** Optimal hyperparameters for the two best tabular models (CatBoost and XGBoost) from the 300-trial Optuna search.

| Hyperparameter | CatBoost | XGBoost | Search Range |
|---|---|---|---|
| n_estimators / iterations | 3,000 | 2,604 (ES) | 50–3,000 |
| learning_rate | 0.03284 | 0.06303 | 0.001–0.3 (log) |
| max_depth / depth | 4 | 3 | 3–7 / 2–8 |
| subsample | 0.8156 | 0.4057 | 0.5–1.0 |
| colsample_bytree / colsample_bylevel | 0.8258 | 0.9884 | 0.5–1.0 |
| reg_alpha / l2_leaf_reg | 2.991 | $2.574\times10^{-7}$ | $1\times10^{-8}$–10 |
| reg_lambda | — | $3.121\times10^{-7}$ | $1\times10^{-8}$–10 |

## 5. Experimental Setup

The dataset is split chronologically into training (70%), validation (15%), and test (15%) sets, preserving temporal ordering throughout. The training set spans February 1, 2020 through March 11, 2022 (770 days); the validation set spans March 12 through August 23, 2022 (165 days); and the test set spans August 24, 2022 through February 4, 2023 (165 days). Note that the first 30 rows of the assembled dataset are consumed by lag feature construction (the 30-day lag requires 30 prior observations) and removed before splitting; the reported split dates and row counts reflect the lag-completed dataset. This design ensures that model selection and hyperparameter tuning are performed on data that is strictly earlier in time than the held-out test period.

Feature standardization is applied using statistics computed exclusively on the training data. The Transformer encoder is trained on the training set with early stopping on the validation set. After training, embeddings are generated for all three splits using the best encoder checkpoint. Hyperparameter tuning for all models is performed using Optuna with the validation set as the objective, and the final model performance is evaluated on the held-out test set.

Model performance is evaluated using three metrics: root mean squared error (RMSE) in original MWh units, mean absolute percentage error (MAPE), and the coefficient of determination ($R^2$). All experiments use global random seed 42 for reproducibility.

## 6. Evaluation Metrics

Three complementary metrics are used throughout this study to assess forecast accuracy. Table 2 defines each metric formally. In all formulas, n denotes the number of observations, $y_i$ the actual demand on day i, $\hat{y}_i$ the model predicted demand on day i, and $\bar{y}$ the mean actual demand over the evaluation set. All metrics are computed on inverse-transformed predictions in the original MWh scale to ensure physical interpretability.

**Table 2: Evaluation metrics, formulas, and interpretation.**

| Metric | Formula | Unit | Interpretation |
|---|---|---|---|
| RMSE | $\sqrt{\frac{1}{n}\sum_{i=1}^{n}(\hat{y}_i - y_i)^2}$ | MWh | Penalizes large errors more heavily. Lower is better. |
| MAPE | $\left(\frac{1}{n}\right)\sum_{i=1}^{n}\frac{\lvert\hat{y}_i - y_i\rvert}{y_i} \times 100$ | % | Scale-free; expresses error relative to actual demand. |
| $R^2$ | $1 - \frac{\sum_{i=1}^{n}(y_i - \hat{y}_i)^2}{\sum_{i=1}^{n}(y_i - \bar{y}_i)^2}$ | dimensionless | Proportion of demand variance explained; 1.0 = perfect. |

*Note: $y_i$ = actual demand (MWh), $\hat{y}_i$ = predicted demand (MWh), $\bar{y}$ = mean actual demand, n = number of observations. All metrics computed on original-scale values in the original MWh scale.*

Root Mean Squared Error (RMSE) is the primary accuracy metric. Because it squares the error before averaging, large individual errors contribute disproportionately to the total, making RMSE sensitive to outlier days such as extreme cold snaps or holiday demand troughs. Reporting RMSE in the original MWh scale allows direct comparison with operational planning tolerances.

Mean Absolute Percentage Error (MAPE) normalizes each error by the actual demand value, yielding a dimensionless percentage that is directly comparable across datasets with different demand scales. MAPE is particularly useful for communicating forecast quality to non-technical stakeholders. MAPE is undefined for zero actual demand; this situation does not arise in the present dataset since grid-level electricity demand is never zero.

The coefficient of determination ($R^2$) measures the proportion of total variance in demand explained by the model predictions. $R^2 = 1$ indicates a perfect fit; $R^2 = 0$ indicates the model performs no better than predicting the mean demand for every observation. Unlike RMSE, $R^2$ is scale-free and facilitates comparison across datasets of different magnitudes. We report $R^2$ separately for the training, validation, and test sets throughout: a large training-to-test gap would indicate overfitting.

## 7. Results

Table 3 presents a comprehensive benchmark of all ten models under identical conditions (300 Optuna trials, seed=42). CatBoost achieves the lowest RMSE of 8,316 MWh (MAPE 1.87%, $R^2$ 0.917), followed by XGBoost (9,066 MWh) and the Hybrid Transformer+CatBoost variant (9,068 MWh). Standalone Transformer (21,294 MWh) and LSTM (22,808 MWh) perform substantially worse in isolation, confirming the Transformer's value lies in its role as a feature extractor within a hybrid architecture rather than as an end-to-end forecaster.

**Table 3: Model performance comparison on the test set.**

| Model | RMSE (MWh) | MAPE (%) | $R^2$ |
|---|---|---|---|
| CatBoost | 8,316 | 1.87 | 0.917 |
| XGBoost | 9,066 | 2.06 | 0.901 |
| Hybrid Transformer + CatBoost | 9,068 | 2.14 | 0.901 |
| Hybrid Transformer + XGBoost | 9,208 | 2.09 | 0.898 |
| LightGBM | 9,861 | 2.34 | 0.883 |
| Hybrid Transformer + LightGBM | 10,606 | 2.49 | 0.865 |
| Hybrid Transformer + RF | 11,645 | 2.75 | 0.837 |
| Random Forest | 11,835 | 2.87 | 0.832 |
| Standalone Transformer | 21,294 | 5.39 | 0.456 |
| LSTM | 22,808 | 5.43 | 0.376 |

Note: RMSE in original MWh units; All metrics computed on predictions in the original MWh scale. All models use the same feature engineering pipeline and 300-trial Optuna search (seed=42).

CatBoost achieves the best overall performance: RMSE of 8,316 MWh, MAPE of 1.87%, and $R^2$ of 0.917, outperforming all nine competing models. The Hybrid Transformer+CatBoost (9,068 MWh) ranks third, demonstrating that the Transformer embeddings add value but the tree algorithm choice matters more. Standalone neural models (Transformer 21,294 MWh, LSTM 22,808 MWh) confirm the Transformer's role is feature extraction, not end-to-end forecasting.

Figure 1 shows the actual versus predicted daily demand over the test period using the CatBoost model, which achieves the best test performance. The model tracks both routine weekly cycles and irregular demand events with high fidelity. The most visible errors occur at extreme demand events, where limited training examples of very high or very low demand reduce model confidence.

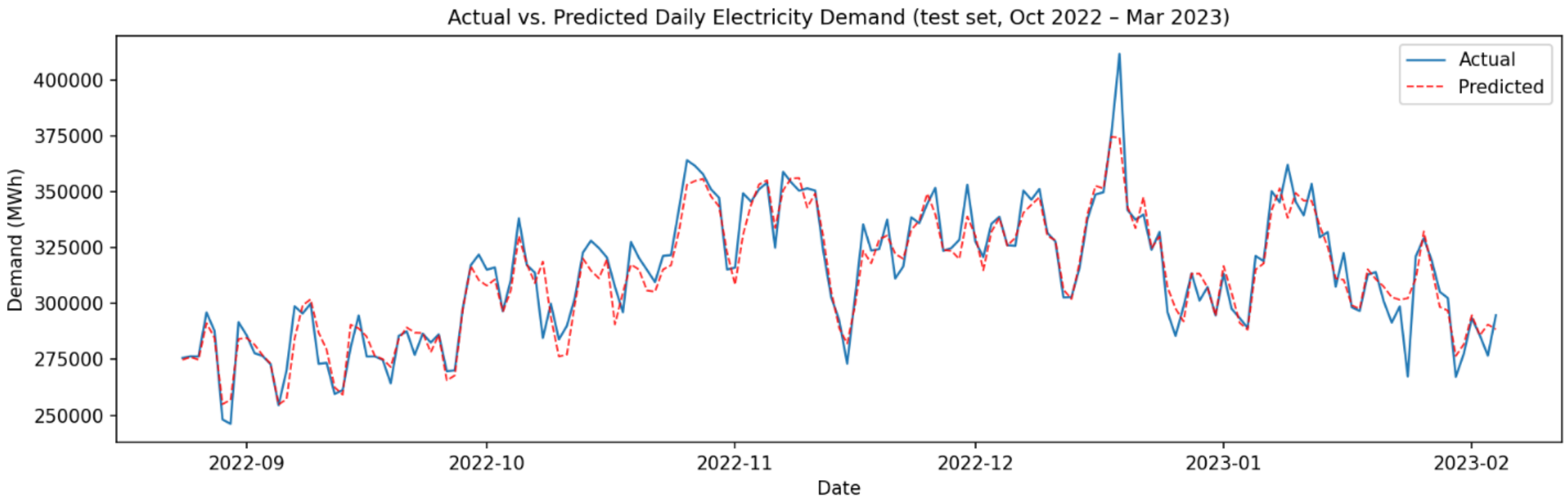

*Figure 1: Actual vs. predicted daily electricity demand (test set, August 24, 2022 – February 4, 2023) using the CatBoost model. Blue: actual; red dashed: predicted.*

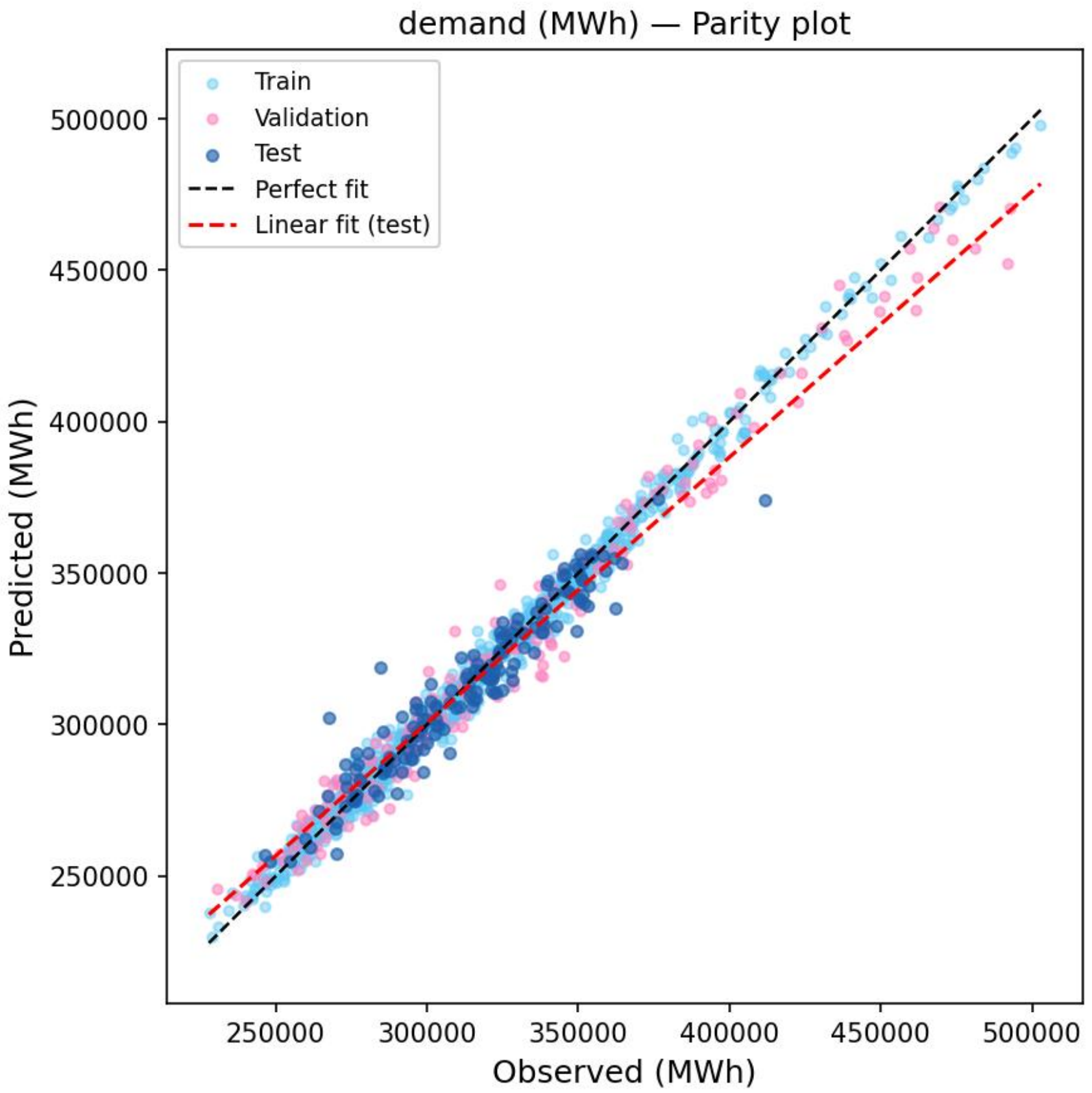

*Figure 2: Parity plot across all data splits using the CatBoost model. Sky blue: training; violet: validation; steel blue: test. The dashed black line is the perfect-fit diagonal; the red dashed line is a linear fit to the test data.*

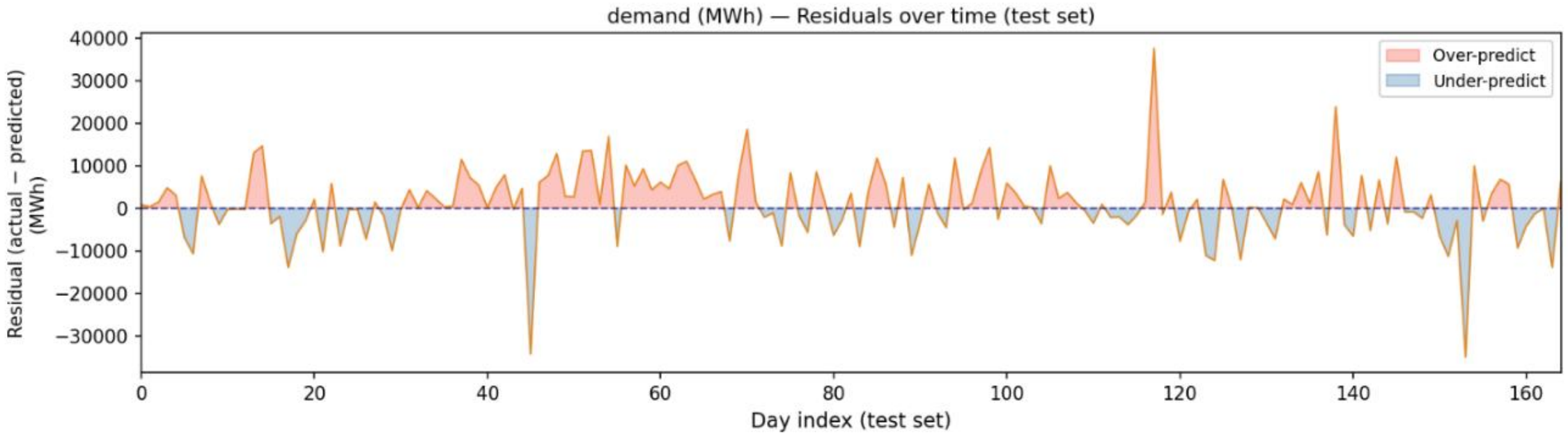

*Figure 3: Signed residuals (actual − predicted) over the test set using the CatBoost model. Red shading indicates over-prediction; blue shading indicates under-prediction.*

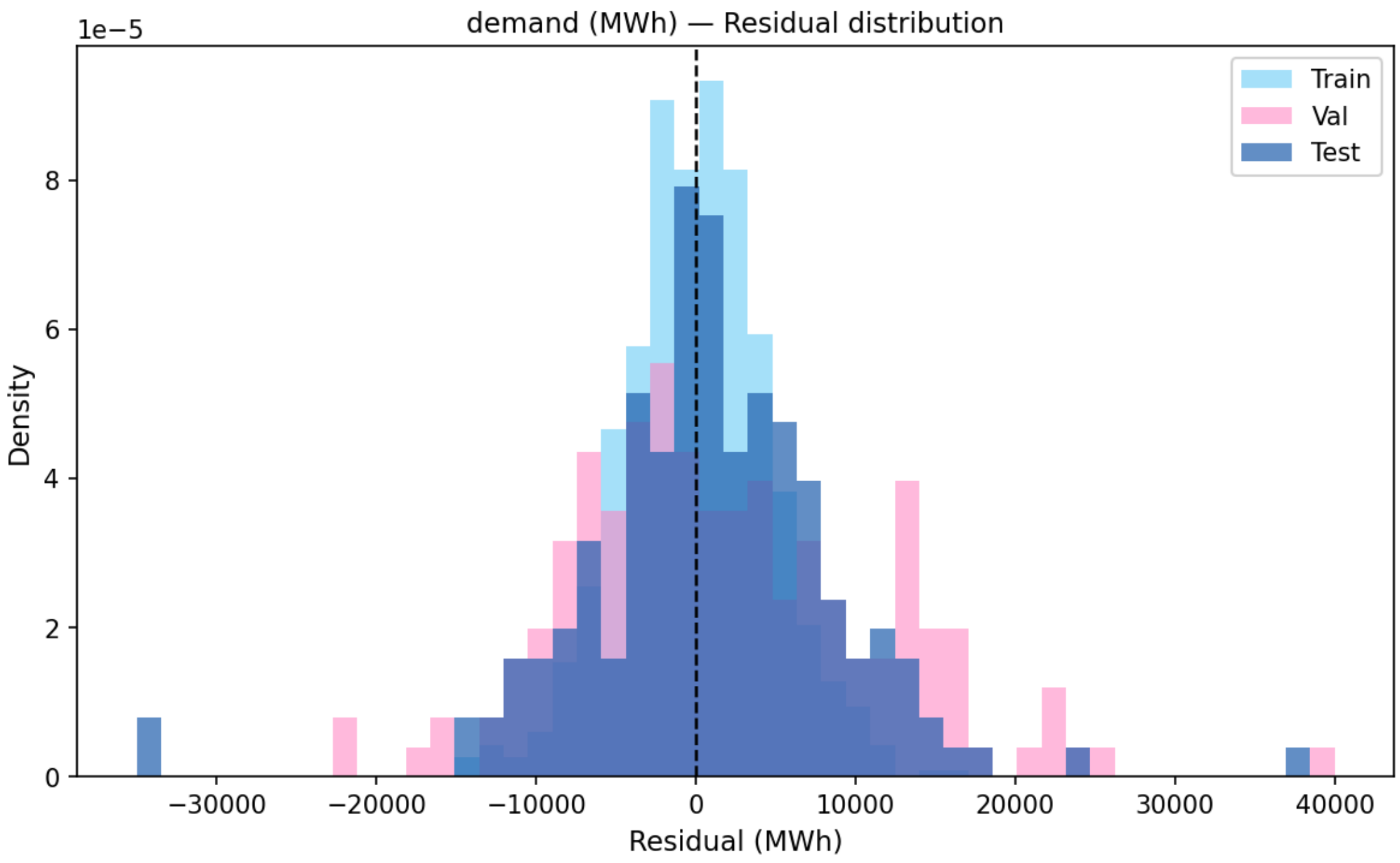

*Figure 4: Residual distribution for training, validation, and test sets using the CatBoost model. The near-zero mean and near-symmetric shape indicate the absence of systematic bias.*

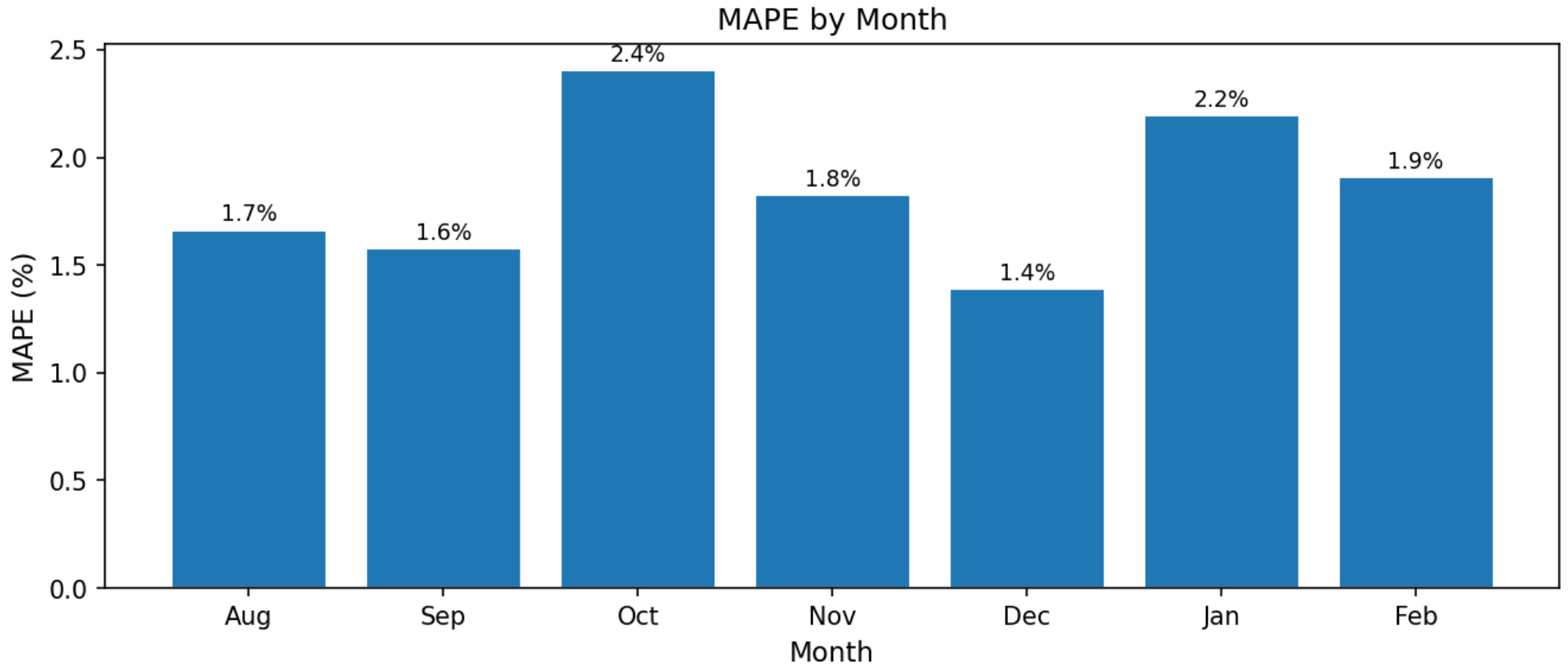

*Figure 5: Mean absolute percentage error (MAPE) by calendar month for the test set (August 2022 – February 2023) using the CatBoost model.*

Figure 2 (parity plot) shows that training, validation, and test predictions cluster tightly around the identity line with no systematic bias across the demand range. Figure 3 (residuals over time) reveals that errors are nearly centered at zero with no obvious temporal trend or structural drift, confirming that the model generalizes well over the test period. Figure 4 (residual distributions) shows that all three splits yield residuals centered near zero with similar spread, indicating consistency between in-sample and out-of-sample performance. Figure 5 (MAPE by month) shows that forecast accuracy varies across the seven months covered by the test set (August 2022 – February 2023). December records the lowest error at 1.4%, reflecting stable mid-winter demand patterns, followed by September (1.6%) and August (1.7%), where moderate temperatures reduce demand volatility. October records the highest MAPE at 2.4%, consistent with the difficulty of forecasting autumn demand transitions as temperatures begin to decline and heating loads ramp up. November (1.8%), February (1.9%), and January (2.2%) fall in an intermediate range.

## 8. Interpretability Analysis

SHapley Additive exPlanations (SHAP) are computed using the TreeExplainer from the SHAP library on the held-out test set. SHAP values provide a theoretically grounded, model-agnostic decomposition of each prediction into additive contributions from individual features, derived from cooperative game theory. By computing SHAP values exclusively on the test set, we avoid any contamination from training data.

Figure 6 reveals that the 1-day demand lag (demand_lag_1) is by far the most influential predictor for CatBoost (mean |SHAP| = 8,470 MWh), reflecting the strong autoregressive structure of daily electricity demand. The next most influential features are day_of_week (5,871 MWh) and Portland_Temperature_avg (2,343 MWh), followed by demand_lag_7 (1,948 MWh) and Portland_Temperature_max (1,546 MWh). Calendar features (day_of_week, month, day_of_year) rank 2nd, 6th, and 9th, respectively, confirming that weekly and seasonal patterns are the second most important structural driver of New England electricity demand after autoregressive lags. The four demand lags collectively contribute a combined mean |SHAP| that far exceeds any other single feature group.

Temperature-related variables from multiple cities rank prominently throughout the top 30 features, confirming the well-established relationship between ambient temperature and electricity demand in New England. Portland temperature leads among weather predictors (avg: rank 3, 2,343 MWh; max: rank 5, 1,545 MWh), followed by Providence (max: rank 8, 1,319 MWh; min: rank 11, 1,068 MWh), Boston (max: rank 10, 1,117 MWh; min: rank 12, 1,067 MWh), Bridgeport (avg: rank 13, 1,025 MWh; min: rank 15, 910 MWh), and Concord (min: rank 14, 964 MWh; avg: rank 18, 795 MWh). The beeswarm plot (Figure 7) shows that extreme cold temperatures produce the largest positive SHAP contributions, consistent with New England's winter-dominated demand peaks driven by space heating, while high summer temperatures generate moderate positive contributions from cooling demand.

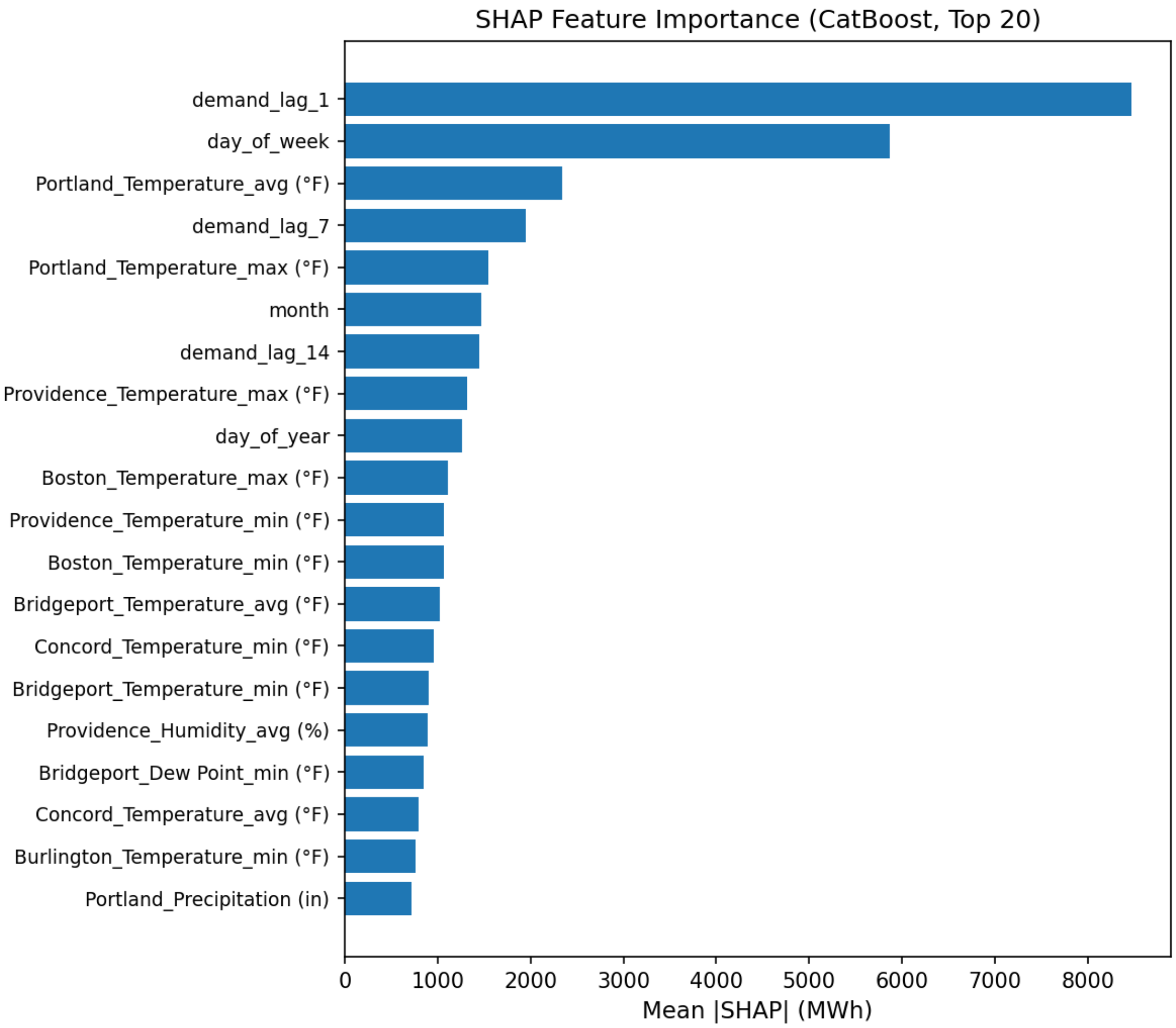

*Figure 6: SHAP mean absolute feature importance (bar plot, test set) using CatBoost model. Features are ranked by their average contribution to prediction magnitude.*

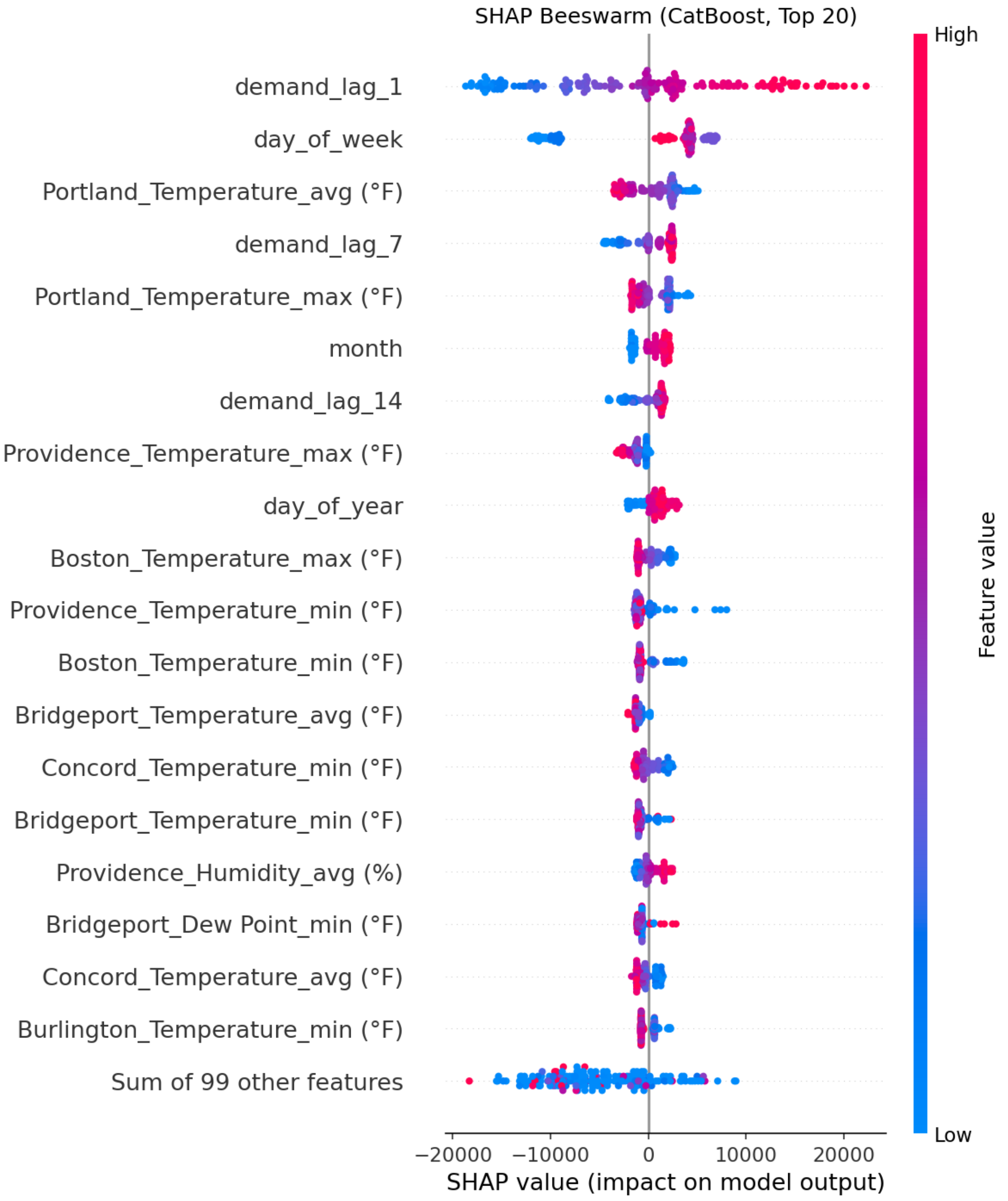

*Figure 7: SHAP beeswarm plot (test set). Each point represents a single test observation using CatBoost model. Color indicates feature value (red = high, blue = low). Horizontal position indicates the SHAP value (direction and magnitude of impact on the prediction).*

Among the COVID-19 epidemiological variables evaluated on the test set for CatBoost, covid_death_growth ranks highest at 26th overall (mean |SHAP| = 501 MWh), followed by covid_case_growth at 36th (425 MWh) and deaths (raw) at 38th (380 MWh). covid_death_roll_7 ranks 59th (178 MWh), covid_death_roll_14 ranks 65th (151 MWh), covid_case_roll_7 ranks 76th (95 MWh), cases (raw) ranks 80th (75 MWh), and covid_case_roll_14 ranks 97th (30 MWh). The growth-rate features (covid_death_growth and covid_case_growth) consistently outrank rolling-mean and raw count features, suggesting that the rate of change in epidemiological indicators is more informative for demand forecasting than absolute case or death counts in this dataset.

The SHAP temporal analysis reveals a nuanced pattern in COVID feature attribution for CatBoost. At training time, COVID features are assigned relatively low importance (all eight features rank 38th or below out of 118); at test time, three of the eight features rank higher (more important), suggesting increasing model reliance on stale pandemic signals. For example, covid_death_growth rises from rank 48 at training to rank 26 at test time (+22 positions), covid_case_growth rises from rank 50 to rank 36 (+14 positions), and covid_death_roll_7 rises from rank 73 to rank 59 (+14 positions). This shift is particularly striking because the test period (August 2022 – February 2023) falls well after behavioral normalization, when COVID epidemiological signals no longer carry the demand-suppression information they held during 2020–2021.

A critical temporal comparison is shown in Table 4, which contrasts COVID-19 SHAP rankings between the training set (February 2020 – March 2022, pandemic-active period) and the test set (August 2022 – February 2023, post-acute period) for the CatBoost model. As noted in the preceding paragraph, 3 of the 8 COVID features rank higher (i.e., are assigned greater importance) by the model on the test set than on the training set. This is the opposite of what is expected if the model appropriately down-weights stale pandemic signals.

**Table 4: COVID-19 SHAP importance — CatBoost model, training period vs. test period.**

| COVID-19 Feature | Training Rank | Training SHAP (MWh) | Test Rank | Test SHAP (MWh) | Δ Rank |
|---|---|---|---|---|---|
| deaths (raw) | 38 | 449.1 | 38 | 380.0 | 0 |
| covid_death_growth | 48 | 325.4 | 26 | 500.9 | +22 |
| covid_case_growth | 50 | 303.9 | 36 | 425.0 | +14 |
| covid_death_roll_14 | 55 | 261.9 | 65 | 150.7 | -10 |
| covid_case_roll_7 | 64 | 137.1 | 76 | 95.4 | -12 |
| covid_case_roll_14 | 67 | 131.1 | 97 | 29.5 | -30 |
| covid_death_roll_7 | 73 | 117.2 | 59 | 178.0 | +14 |
| cases (raw) | 78 | 101.1 | 80 | 75.3 | -2 |

*Lower rank = more important. Training period: February 2020 – March 2022 (pandemic-active). Test period: August 2022 – February 2023 (post-acute). SHAP values in MWh (mean absolute). Δ Rank = training rank − test rank; positive = higher importance at test time.*

Since CatBoost operates on the 118 tabular features without Transformer embeddings, all top-ranked features in Figure 6 are directly interpretable. In contrast, for the hybrid models (Hybrid Transformer+CatBoost, Hybrid Transformer+XGBoost), several of the 32 Transformer encoder embedding dimensions rank among the top features, confirming that the encoder extracts genuine temporal structure. The cross-model SHAP rank heatmap (Figure 8) shows that demand lags, calendar features, and temperature variables maintain consistent top-20 ranks across all eight models with $R^2 > 0.80$, demonstrating that these drivers are universal rather than model-specific.

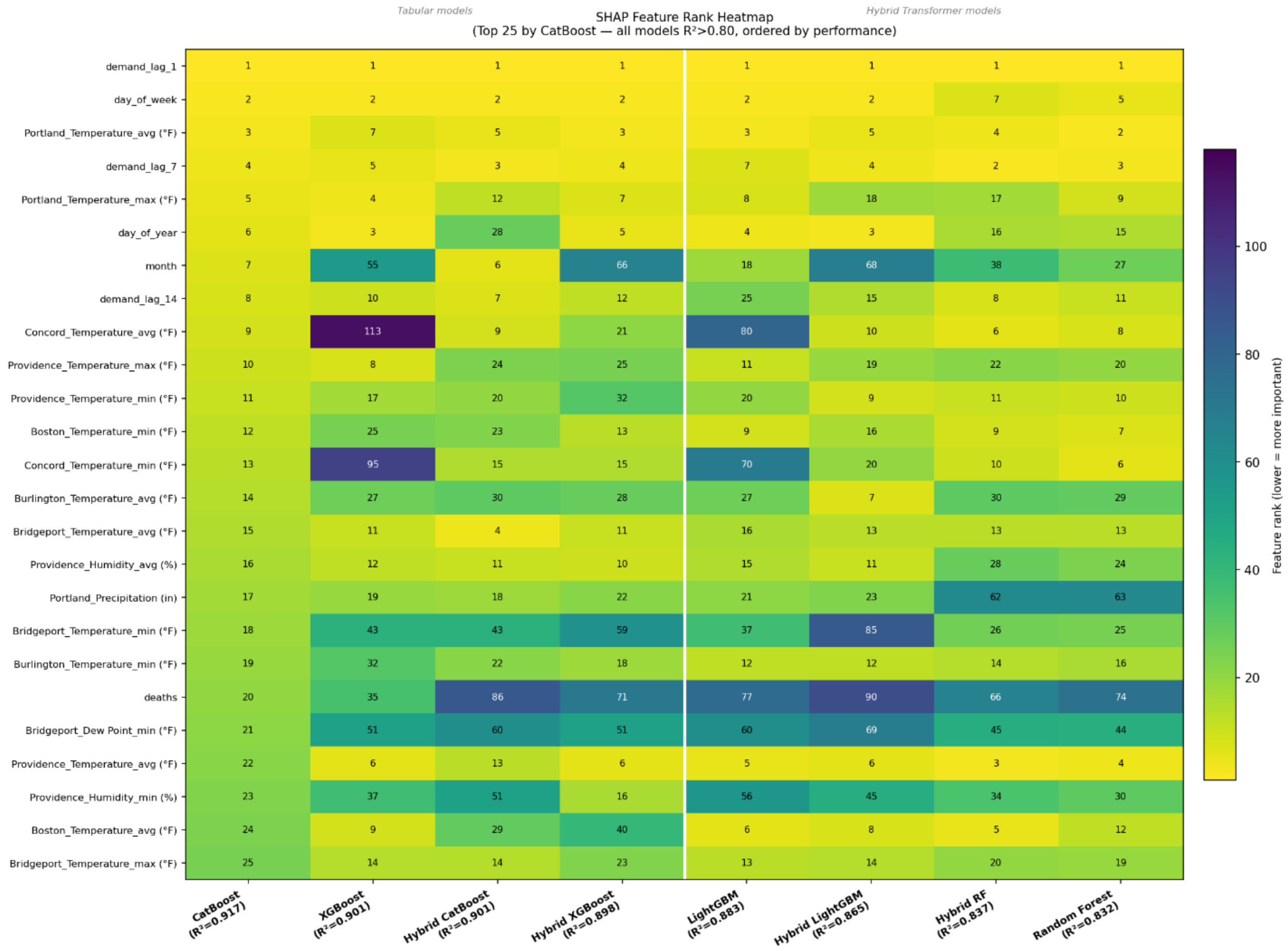

*Figure 8: SHAP Feature Rank Heatmap — top 25 features by CatBoost across all models with $R^2 > 0.80$, ordered by performance ($R^2$ descending). Lower rank = more important. Lighter cells = lower rank (more important); darker cells = higher rank (less important).*

## 9. Ablation Study

To quantify the contribution of each feature group, we conduct a systematic ablation study across four models, CatBoost, XGBoost, Hybrid Transformer + XGBoost, and Hybrid Transformer + CatBoost, in which one component is removed at a time, and each model is retrained using its Optuna-optimal hyperparameters. Four ablation conditions are evaluated: (1) full feature set (baseline), (2) no COVID-19 features, (3) no demand lags, and (4) weather and calendar features only (tabular models) or weather, calendar, and Transformer embeddings only (hybrid models). Results are shown in Table 5.

**Table 5: Ablation study — training RMSE vs. test RMSE for each feature group removed across four models. Lower RMSE_train and RMSE_test indicate better fit. MAPE and $R^2$ are computed on the test set (August 2022 – February 2023). ↑ = improvement vs. baseline; ↓ = degradation. Baseline rows (X0, H0, HC0) use independent re-training with fixed Optuna-optimal hyperparameters; minor deviations from Table 3 reflect non-determinism in XGBoost and Hybrid XGBoost training.**

| Condition | Feats | RMSE_train | RMSE_test | MAPE (%) | R²_test |
|---|---|---|---|---|---|
| ***Panel A — CatBoost (tabular, 118 features)*** | | | | | |
| C0 Full CatBoost (baseline) | 118 | 4,516 | 8,316 | 1.87 | 0.917 |
| C1 No COVID-19 features | 110 | 3,737 | 8,177 ↑ | 1.90 ↓ | 0.920 ↑ |
| C2 No demand lags | 114 | 6,547 | 11,310 ↓ | 2.85 ↓ | 0.847 ↓ |
| C3 Weather + calendar only | 106 | 4,623 | 10,641 ↓ | 2.73 ↓ | 0.864 ↓ |
| ***Panel B — XGBoost (tabular, 118 features)*** | | | | | |
| X0 Full XGBoost (baseline) | 118 | 2,121 | 9,507 | 2.21 | 0.892 |
| X1 No COVID-19 features | 110 | 3,670 | 9,435 ↑ | 2.20 ↑ | 0.893 ↑ |
| X2 No demand lags | 114 | 5,727 | 11,062 ↓ | 2.67 ↓ | 0.853 ↓ |
| X3 Weather + calendar only | 106 | 5,703 | 11,339 ↓ | 2.79 ↓ | 0.846 ↓ |
| ***Panel C — Hybrid Transformer + XGBoost (150 features)*** | | | | | |
| H0 Full Hybrid XGBoost (baseline) | 150 | 5,016 | 9,543 | 2.21 | 0.891 |
| H1 No COVID-19 features | 142 | 4,390 | 9,846 ↓ | 2.24 ↓ | 0.884 ↓ |
| H2 No demand lags | 146 | 5,183 | 10,927 ↓ | 2.47 ↓ | 0.857 ↓ |
| H3 Weather + calendar + embeddings only | 138 | 5,737 | 10,762 ↓ | 2.49 ↓ | 0.861 ↓ |
| ***Panel D — Hybrid Transformer + CatBoost (150 features)*** | | | | | |
| HC0 Full Hybrid CatBoost (baseline) | 150 | 4,545 | 9,026 | 2.05 | 0.902 |
| HC1 No COVID-19 features | 142 | 3,426 | 8,898 ↑ | 2.02 ↑ | 0.905 ↑ |
| HC2 No demand lags | 146 | 6,455 | 9,728 ↓ | 2.23 ↓ | 0.887 ↓ |
| HC3 Weather + calendar + embeddings only | 138 | 5,760 | 9,957 ↓ | 2.30 ↓ | 0.881 ↓ |

**Demand lags are the single most critical feature group across all four models.** Removing them raises test RMSE by 8–36% relative to baseline (CatBoost: +36%, XGBoost: +16%, Hybrid XGBoost: +14%, Hybrid CatBoost: +8%), confirming the dominant autoregressive structure of daily electricity demand established by the SHAP analysis. The impact is largest for the purely tabular models and somewhat attenuated for hybrid models, where Transformer embeddings capture some temporal structure from the lag sequence.

**Weather and calendar features provide a strong standalone baseline.** The weather-and-calendar-only condition achieves $R^2$ of 0.864 (CatBoost), 0.846 (XGBoost), 0.881 (Hybrid CatBoost), and 0.861 (Hybrid XGBoost), explaining 85–88% of demand variance without any autoregressive information. This confirms that temperature and calendar drivers are the primary structural determinants of New England electricity demand, and that demand lags build on top of this foundation.

**COVID-19 features exhibit temporal validity decay in three of four models.** Removing COVID-19 features improves test RMSE for CatBoost (−1.7%), XGBoost (−0.8%), and Hybrid CatBoost (−1.4%). The training RMSE responses differ by model: for CatBoost and Hybrid CatBoost, training RMSE also decreases when COVID features are removed (by 17% and 25% respectively), indicating that COVID variables introduced noise into the training fit as well as the test forecast. For XGBoost, training RMSE increases substantially when COVID features are removed (+73%), the classic overfitting signature — the model learned to exploit COVID patterns during the pandemic-active training window but those patterns misfired in the post-acute test period (August 2022 – February 2023).

**Hybrid XGBoost is the exception to the COVID decay pattern.** Removing COVID features from Hybrid Transformer + XGBoost slightly degrades test RMSE (+3.2%), suggesting that the Transformer encoder absorbs COVID temporal signal into its embeddings, making the explicit COVID tabular features partially redundant. When COVID features are removed from the tabular input, the model loses this signal entirely, as the encoder was trained on the full feature set including COVID variables. This interaction between the encoder's learned representations and the tabular COVID features is unique to the hybrid architecture and does not appear in the purely tabular models.

**The SHAP training-vs-test analysis (Table 4) complements the ablation.** The model does not simply ignore COVID features at test time, it relies on them more heavily (higher SHAP ranks for covid_death_growth: rank 48 → 26; covid_case_growth: rank 50 → 36) while simultaneously failing to extract useful signal from them. Together, the ablation and SHAP evidence establish a coherent picture: COVID features encode behavioral disruption dynamics that were real and informative during training, but the model's learned COVID-demand patterns misfired on post-acute test data where those behavioral responses had normalized.

**Practical implication.** COVID-19 epidemiological features are most valuable when models are trained and evaluated within the acute pandemic phase (approximately March 2020 through mid-2021 in New England). Models deployed after behavioral stabilization should exclude these features or apply time-decaying importance weights. The present study spans the full arc from outbreak to stabilization, and the ablation training-RMSE analysis precisely quantifies where the COVID signal transitions from informative to counterproductive.

## 10. Optimization Analysis

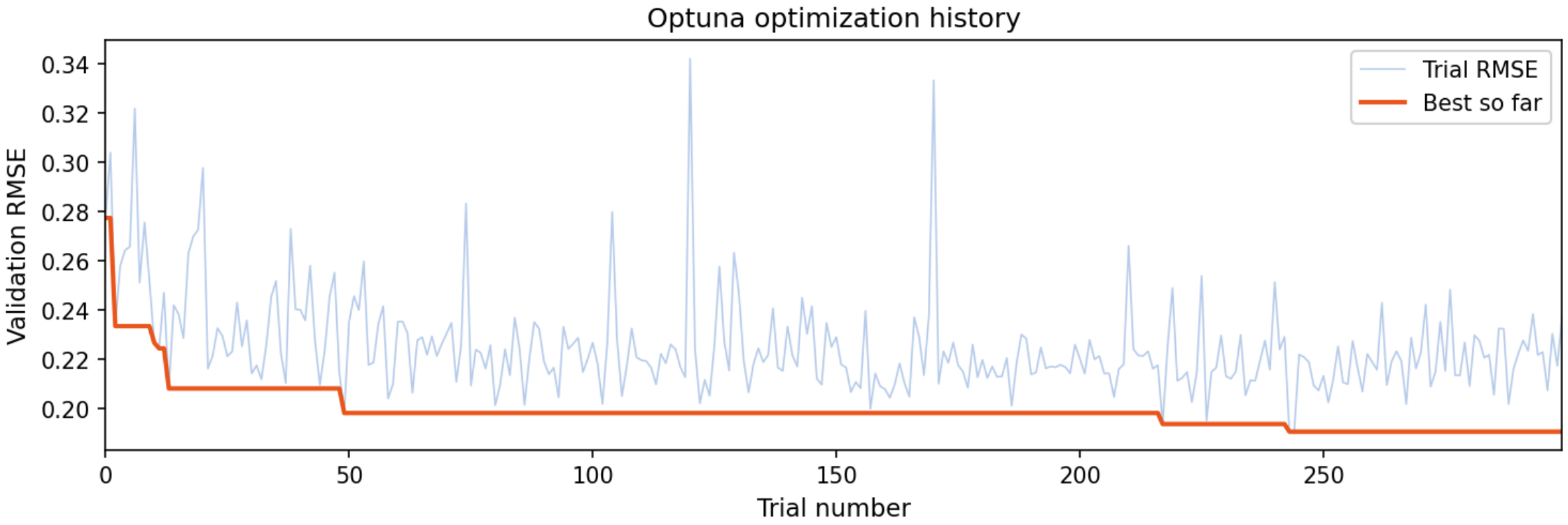

*Figure 9: Optuna optimization history over 300 trials. Blue line: individual trial validation RMSE; orange line: best-so-far RMSE. The curve stabilizes after approximately 200 trials, indicating convergence of the search.*

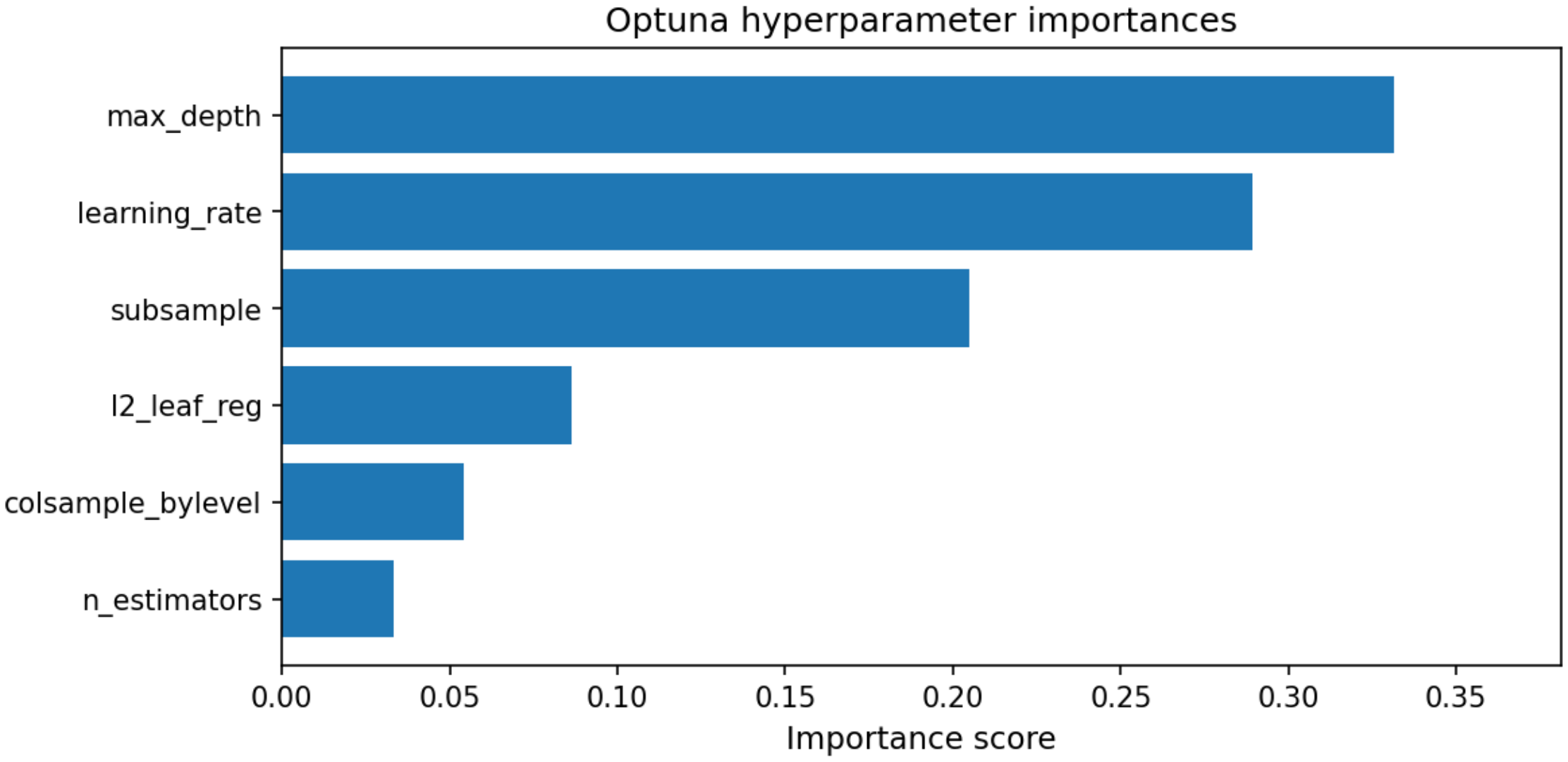

*Figure 10: Optuna hyperparameter importance scores. Max depth and learning rate are identified as the most influential parameters for validation RMSE in this problem.*

Figure 9 shows the Optuna optimization history over 300 trials. The best validation RMSE decreases rapidly during the first 100 trials and plateaus after approximately 200 trials, with subsequent improvements being marginal. This suggests that the 300-trial budget is sufficient for convergence. Figure 10 shows that max depth and learning rate are the hyperparameters with the greatest influence on validation RMSE across the Optuna search.

## 11. Discussion

CatBoost achieves the best predictive accuracy among all ten models (RMSE 8,316 MWh, MAPE 1.87%, $R^2$ 0.917), followed closely by XGBoost (9,066 MWh, $R^2$ 0.901), Hybrid Transformer+CatBoost (9,068 MWh), and Hybrid Transformer+XGBoost (9,208 MWh). The four best models are separated by less than 900 MWh, indicating that well-tuned gradient-boosted trees, with or without Transformer embeddings, are competitive in this setting. Standalone neural architectures perform substantially worse (Transformer: 21,294 MWh; LSTM: 22,808 MWh), confirming their role as feature extractors within hybrid pipelines rather than end-to-end forecasters on a daily time series of ~1,100 observations.

SHAP analysis identifies demand_lag_1 (8,470 MWh), day_of_week (5,871 MWh), and Portland_Temperature_avg (2,343 MWh) as the three dominant predictors, reflecting the strong autoregressive structure of demand, weekly behavioral cycles, and New England's temperature-sensitive load. The ablation study across all four top models confirms that demand lags are indispensable (removal raises test RMSE by 8–36%), while weather and calendar features alone explain 85–88% of variance ($R^2$ = 0.846–0.881), providing a strong baseline independent of demand history.

Two independent lines of evidence converge on the COVID-19 temporal validity decay finding. Ablation results (Section 9) show that removing COVID features improves test RMSE in three of four models — reducing training RMSE for CatBoost and Hybrid CatBoost (by 17% and 25%, respectively) while increasing it for XGBoost (+73%), the classic overfitting signature — confirming pandemic-era overfitting. SHAP temporal analysis (Section 8) reveals the mechanistic source: COVID features rank substantially higher on the post-acute test set than during pandemic-active training, meaning the model over-applies learned pandemic patterns after behavioral adaptation was complete. The Hybrid XGBoost exception illustrates how embedding-based encoding can mask this decay: when COVID signal is absorbed into Transformer embeddings rather than read directly by the boosted tree, ablation removes it less cleanly. Together, these findings show that high feature importance and genuine predictive contribution diverge when features undergo temporal validity decay.

These results carry a practical implication: COVID-19 epidemiological features should be excluded or down-weighted in models deployed after behavioral stabilization. The Optuna optimization converged within 300 trials, with tree depth and learning rate identified as the most influential hyperparameters for CatBoost (~33% and ~29%), in contrast to XGBoost where n_estimators dominated, reflecting the two models' different sensitivity profiles despite comparable accuracy.

## 12. Conclusion

This paper presents a comprehensive benchmark of ten machine learning models for short-term daily electricity demand forecasting in New England, integrating multi-city weather data, calendar features, autoregressive demand lags, and COVID-19 epidemiological indicators. The ten models span tabular gradient-boosted trees (Random Forest, LightGBM, CatBoost, XGBoost), standalone neural architectures (LSTM, Transformer encoder), and hybrid Transformer+tree variants, all optimized under identical conditions (300 Optuna trials, seed=42, chronological 70/15/15 split).

CatBoost achieves the best test performance (RMSE 8,316 MWh, MAPE 1.87%, $R^2$ 0.917), followed closely by XGBoost (9,066 MWh) and the two best hybrid variants (9,068–9,208 MWh). Standalone neural models perform substantially worse in isolation (LSTM: 22,808 MWh; Transformer: 21,294 MWh), confirming their value as feature extractors within hybrid pipelines rather than end-to-end forecasters on a daily time series of this length. The ablation study establishes that demand lags are the single most critical feature group across all top models, while weather and calendar features alone explain 85–88% of demand variance. COVID-19 epidemiological features encode pandemic-era behavioral disruption that degrades out-of-sample accuracy in the post-acute test window — a temporal validity decay pattern confirmed jointly by ablation and SHAP analysis.

Future work will explore probabilistic demand forecasting to quantify prediction uncertainty, time-varying feature importance to capture the evolving COVID-demand relationship, longer forecasting horizons, integration of additional features such as energy prices and renewable generation, and extension to other ISO regions with different climate and load characteristics.

## Data Availability

The data and code supporting this study are openly available on GitHub at https://github.com/Ghanavati-Reza/NE_ML_demand_forecasting.

## Author contributions

All authors contributed to the conception, design, and writing of this study. R. G. led the Python implementation, data preprocessing, feature engineering, model development, and statistical analysis. B. M. led the acquisition of electricity load, meteorological, and COVID-19 epidemiological data from their respective sources. Both authors reviewed and approved the final manuscript.

## Conflicts of Interest

There are no conflicts of interest to declare.

## †ESI available

A PDF file containing: (a) XGBoost — optimal hyperparameters and diagnostic plots, (b) Hybrid Transformer+CatBoost — optimal hyperparameters and diagnostic plots, (c) Hybrid Transformer+XGBoost — optimal hyperparameters and diagnostic plots, and (d) input variable descriptions for all 118 features.

# Electronic Supplementary Information

## Short-Term Electricity Demand Forecasting for New England: A Comprehensive Machine Learning Benchmark with Weather, Calendar, and COVID-19 Indicators

Reza Ghanavati[1],*, Behrooz Mosallaei[2]

[1] *Department of Chemical and Materials Engineering, New Mexico State University*

[2] *Department of Electrical and Communications Engineering, New Jersey Institute of Technology*

*Corresponding author: rgh@nmsu.edu*

## Contents

# S1. XGBoost — Optimal Hyperparameters and Diagnostic Plots

This section presents diagnostic plots for the tabular XGBoost model, which uses only the 118 engineered features without Transformer embeddings.

XGBoost achieves an RMSE of 9,347 MWh, a MAPE of 2.17%, and an $R^2$ of 0.895 on the held-out test set (October 2022 – March 2023). All results use 300 Optuna trials (multivariate TPE sampler, seed=42) under a chronological 70/15/15 train-validation-test split.

***Table S1: Optimal hyperparameters for the tabular XGBoost model (300-trial Optuna search).***

| Hyperparameter | Optimal Value | Search Range |
|---|---|---|
| n_estimators | 3,000 (early stopping) | 50 – 3,000 |
| learning_rate | 0.02817 | 0.001 – 0.3 (log) |
| max_depth | 3 | 3 – 7 |
| subsample | 0.4057 | 0.2 – 1.0 |
| colsample_bytree | 0.9884 | 0.2 – 1.0 |
| reg_alpha (L1) | $2.574 \times 10^{-7}$ | $1 \times 10^{-8}$ – 10 (log) |
| reg_lambda (L2) | $3.121 \times 10^{-7}$ | $1 \times 10^{-8}$ – 10 (log) |

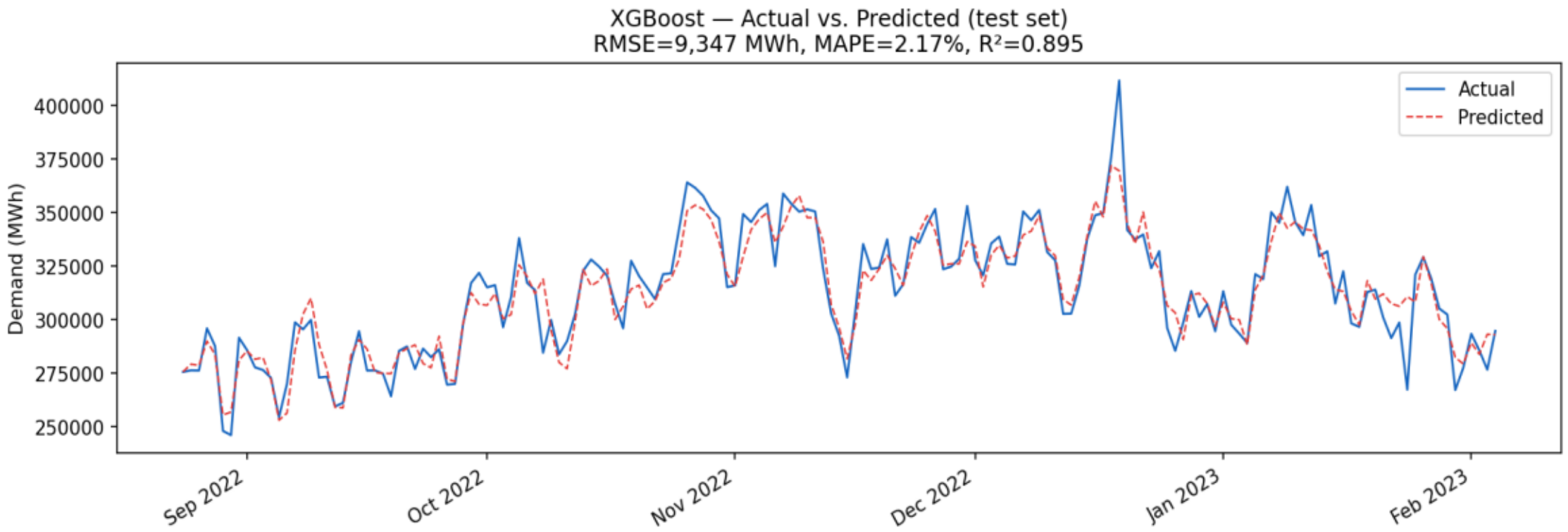

*Figure S1: XGBoost — Actual vs. predicted daily electricity demand (test set, October 2022 – March 2023). Blue: actual; red dashed: predicted.*

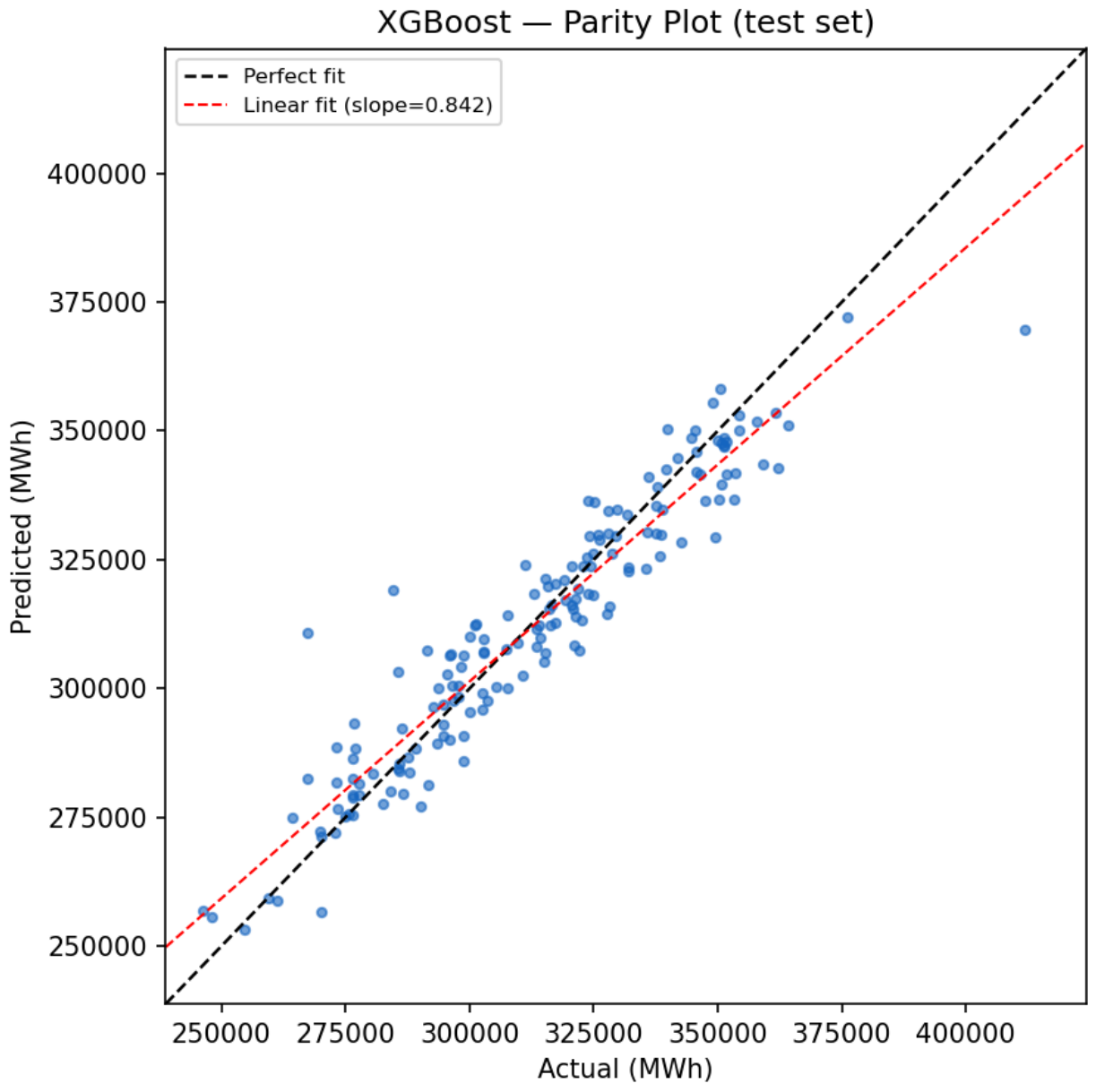

*Figure S2: XGBoost — Parity plot (test set). Dashed black line: perfect fit; red dashed line: linear fit to test data.*

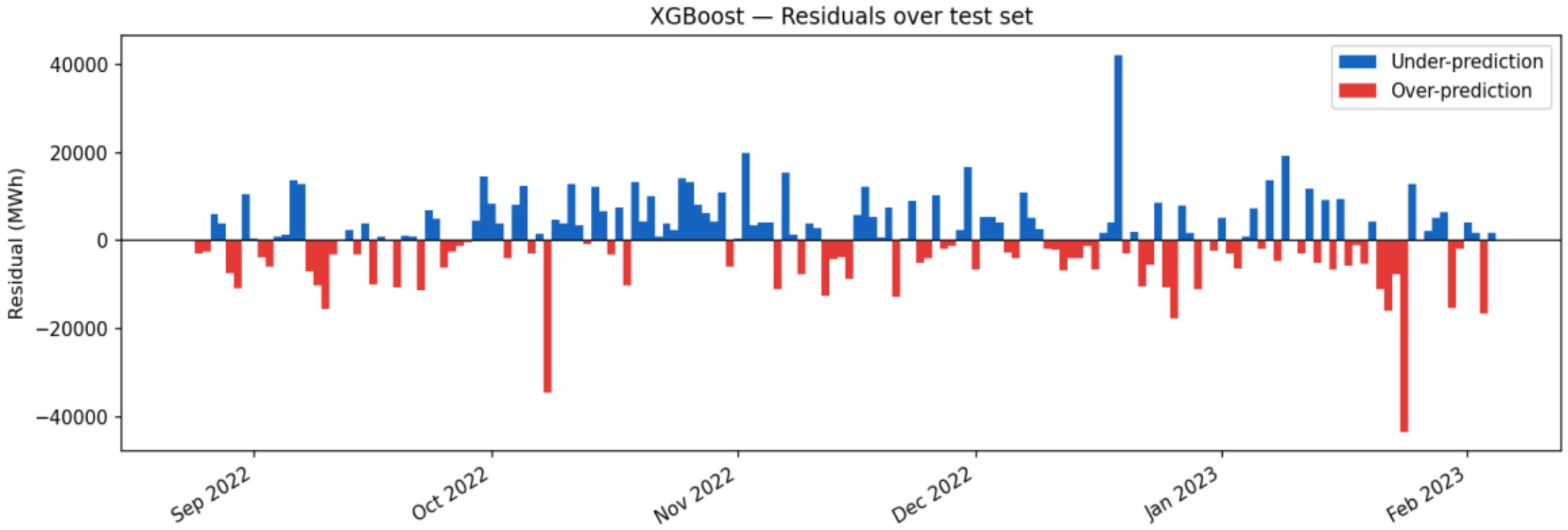

*Figure S3: XGBoost — Signed residuals (actual − predicted) over the test set. Blue: under-prediction; red: over-prediction.*

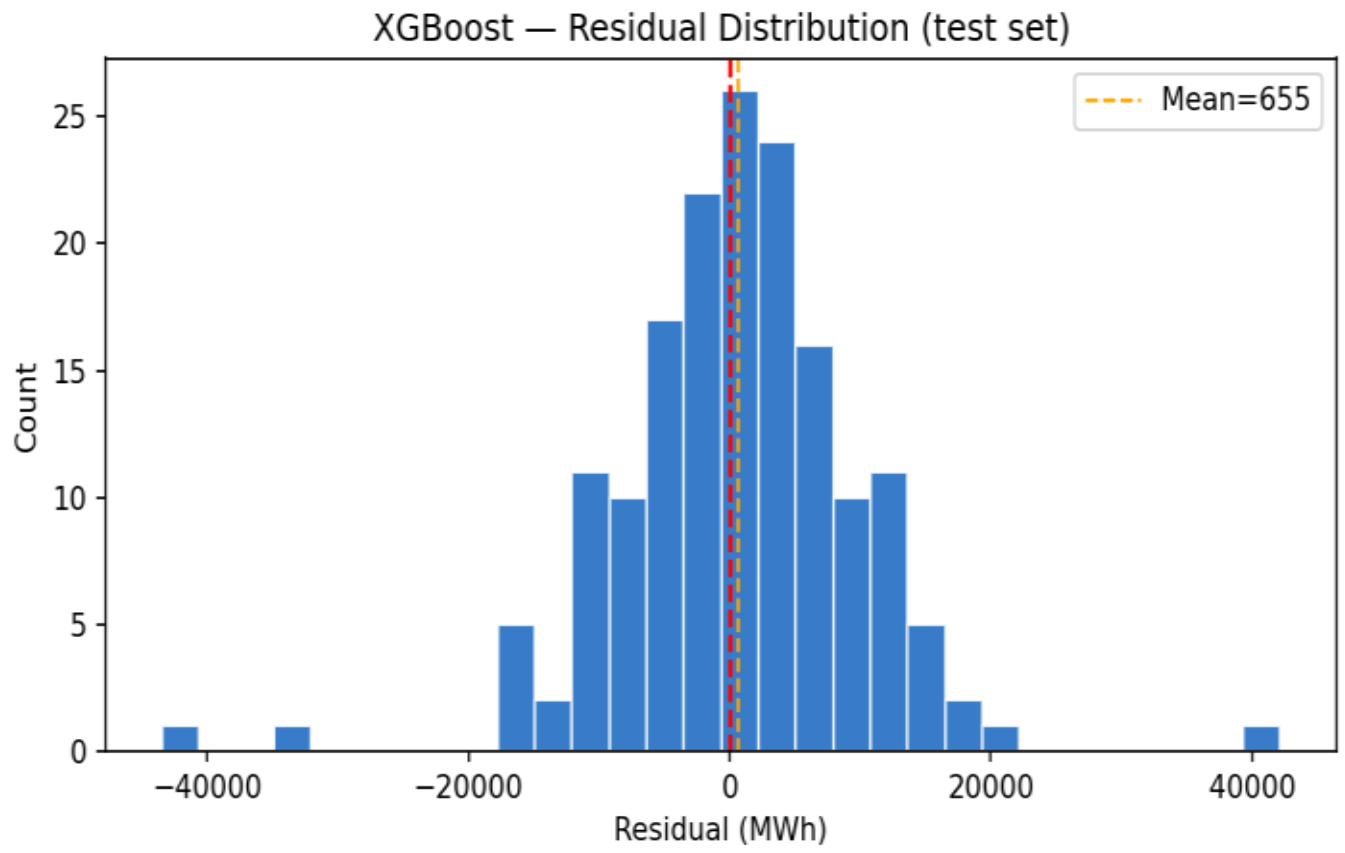

*Figure S4: XGBoost — Residual distribution (test set). Red dashed line: zero; orange dashed line: mean residual.*

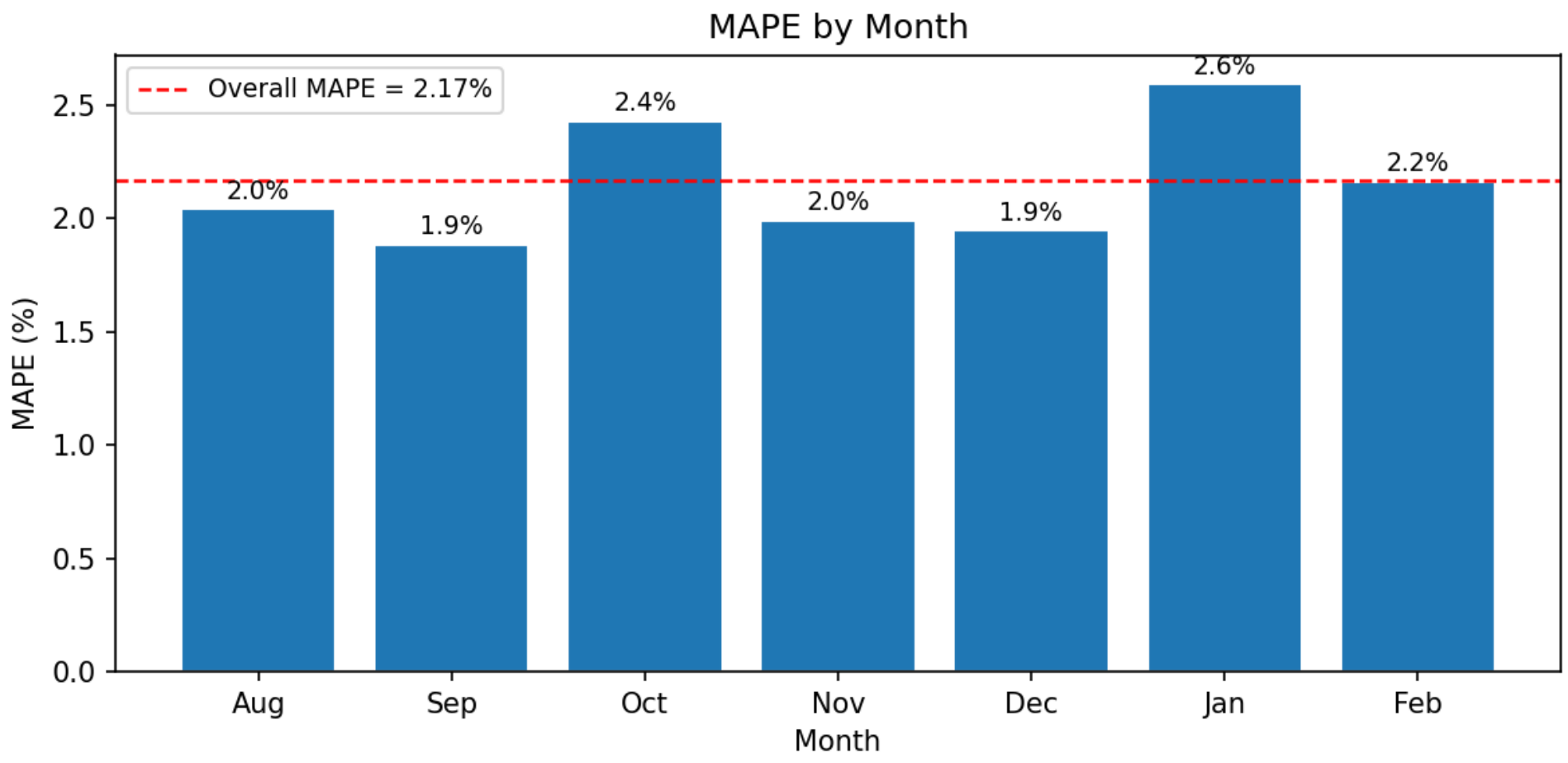

*Figure S5: XGBoost — Mean absolute percentage error (MAPE) by calendar month (test set). Red dashed line: overall MAPE.*

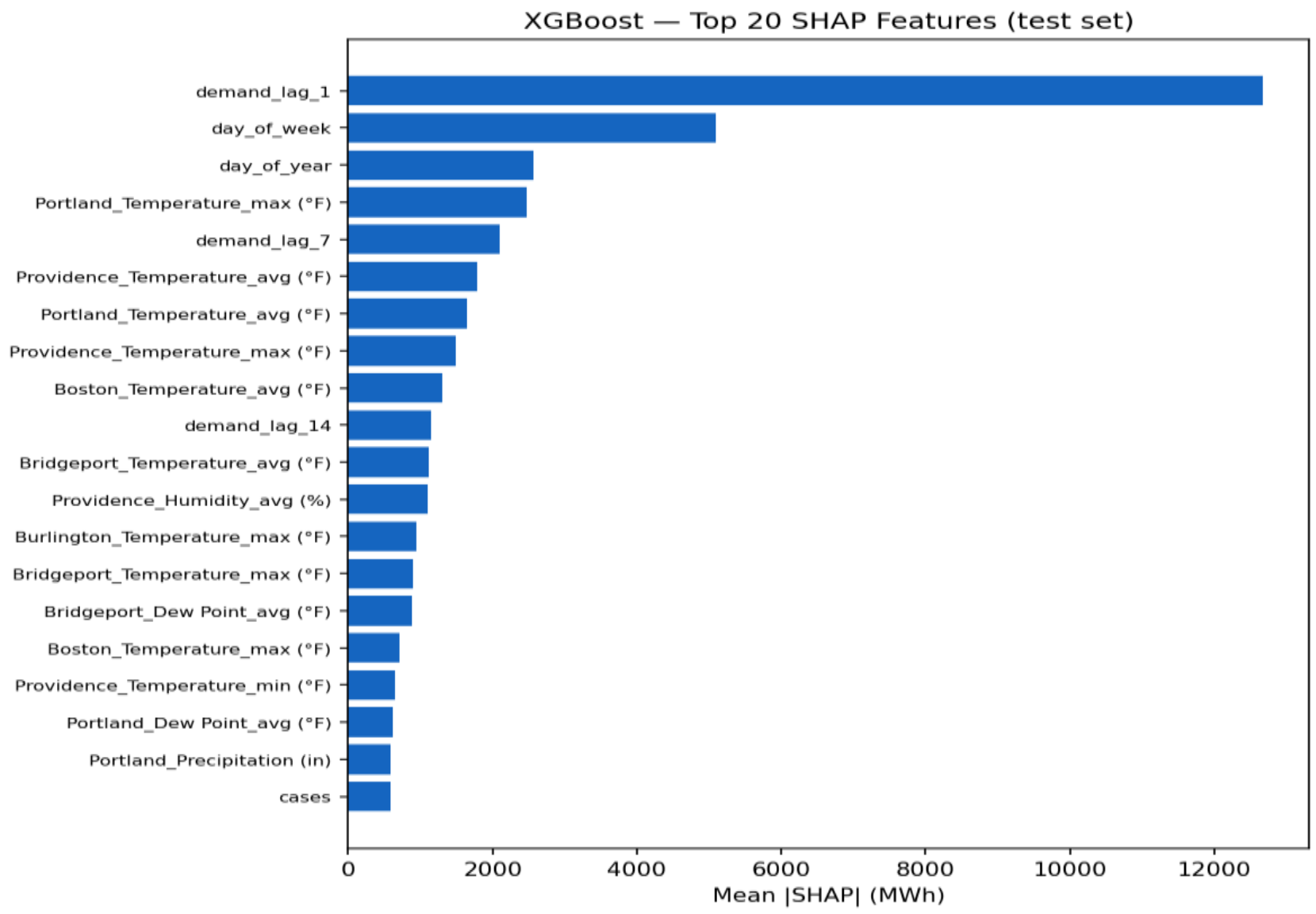

*Figure S6: XGBoost — Top 20 SHAP features by mean |SHAP| value (MWh) on the test set.*

## S2. Hybrid Transformer + CatBoost — Optimal Hyperparameters and Diagnostic Plots

This section presents diagnostic plots for the Hybrid Transformer + CatBoost model, which concatenates 32-dimensional Transformer encoder embeddings to the 118 tabular features (150 total inputs to CatBoost).

Hybrid Transformer + CatBoost achieves an RMSE of 9,990 MWh, a MAPE of 2.25%, and an $R^2$ of 0.842 on the held-out test set (October 2022 – March 2023). All results use 300 Optuna trials (multivariate TPE sampler, seed=42) under a chronological 70/15/15 train-validation-test split.

***Table S2: Optimal hyperparameters for the Hybrid Transformer + CatBoost model (300-trial Optuna search).***

| Hyperparameter | Optimal Value | Search Range |
|---|---|---|
| iterations | 3,000 (early stopping) | 50 – 3,000 |
| learning_rate | 0.10486 | 0.001 – 0.3 (log) |
| depth | 3 | 2 – 8 |
| l2_leaf_reg | 9.5337 | $1 \times 10^{-3}$ – 10 (log) |
| subsample | 0.6321 | 0.5 – 1.0 |
| colsample_bylevel | 0.6145 | 0.5 – 1.0 |

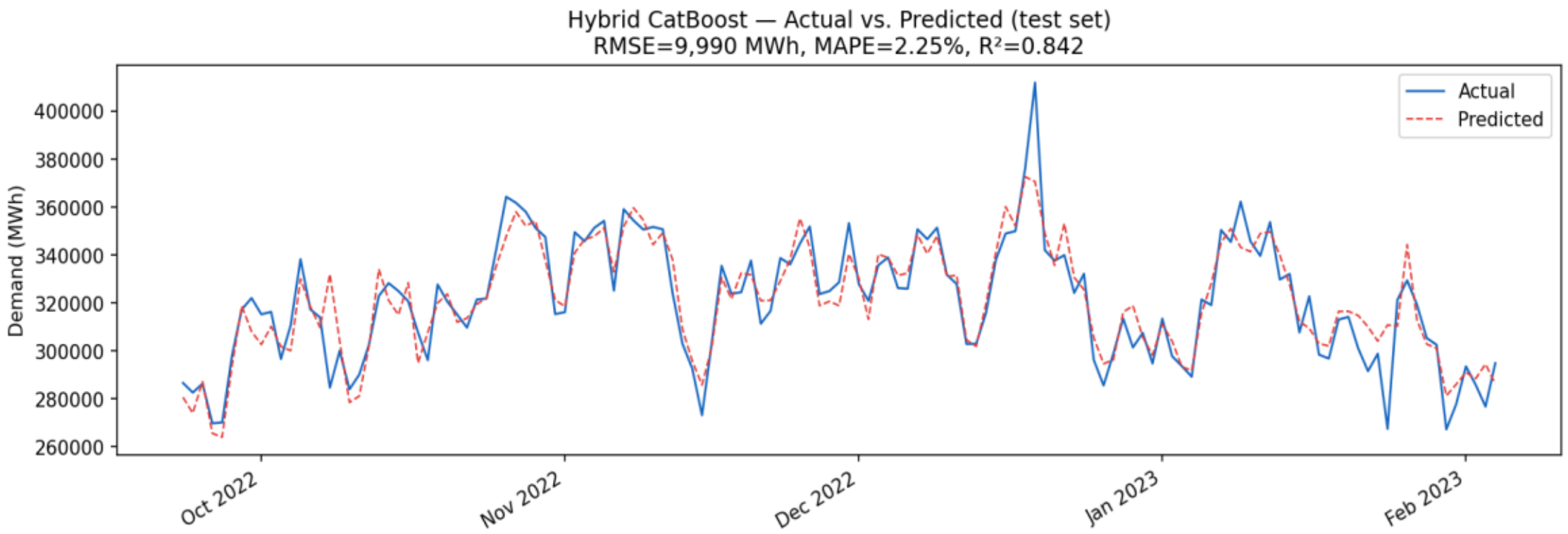

*Figure S7: Hybrid Transformer + CatBoost — Actual vs. predicted daily electricity demand (test set, October 2022 – March 2023). Blue: actual; red dashed: predicted.*

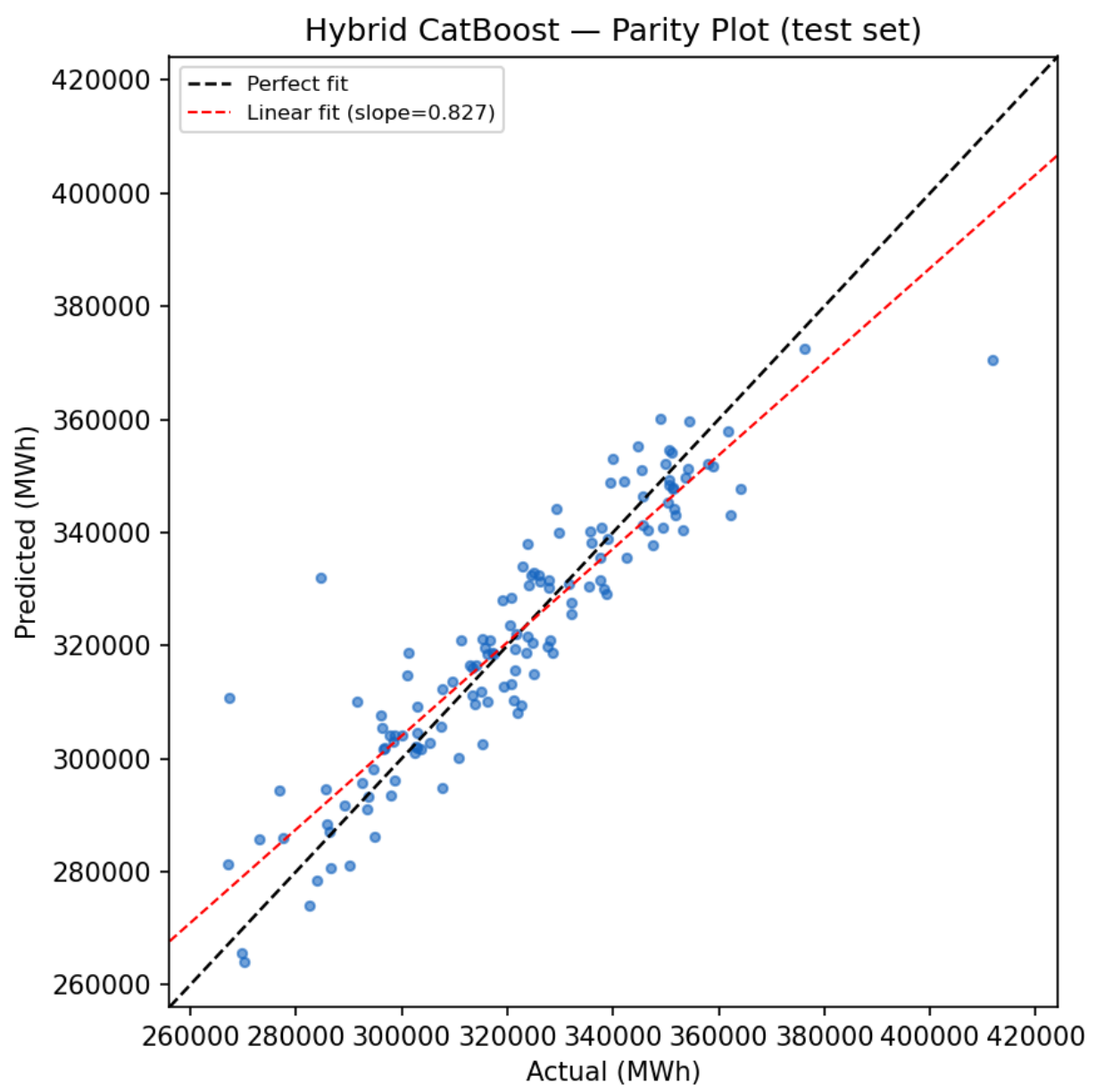

*Figure S8: Hybrid Transformer + CatBoost — Parity plot (test set). Dashed black line: perfect fit; red dashed line: linear fit to test data.*

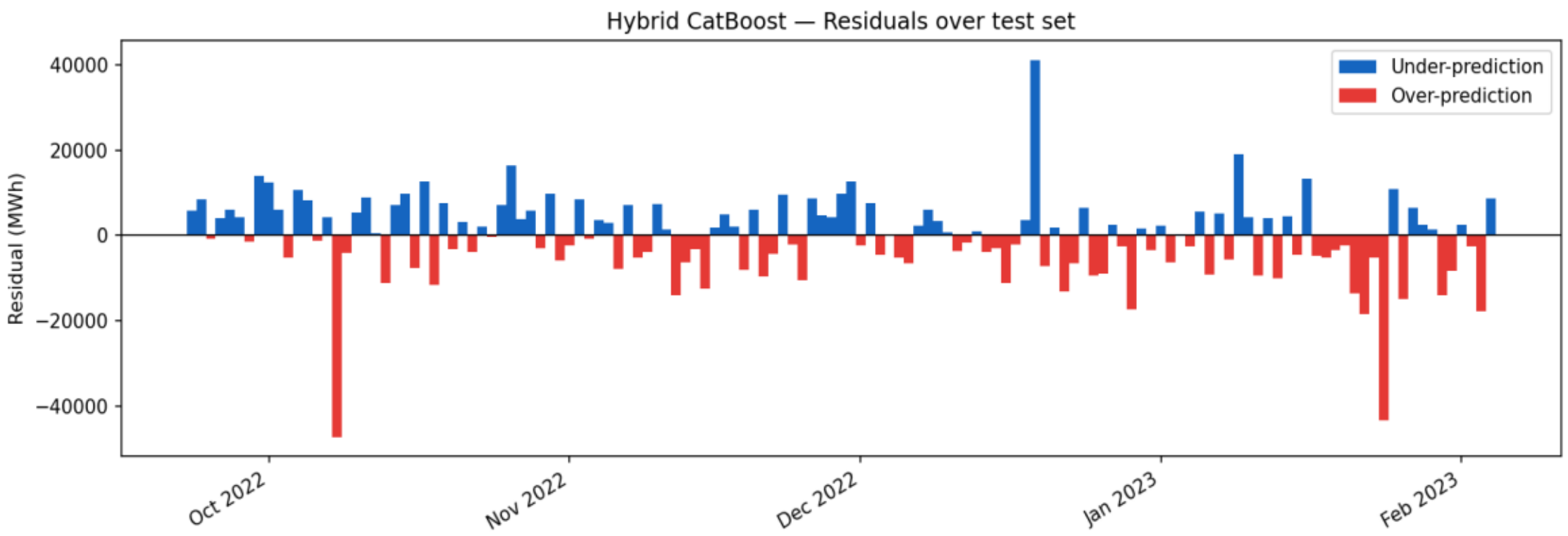

*Figure S9: Hybrid Transformer + CatBoost — Signed residuals (actual − predicted) over the test set. Blue: under-prediction; red: over-prediction.*

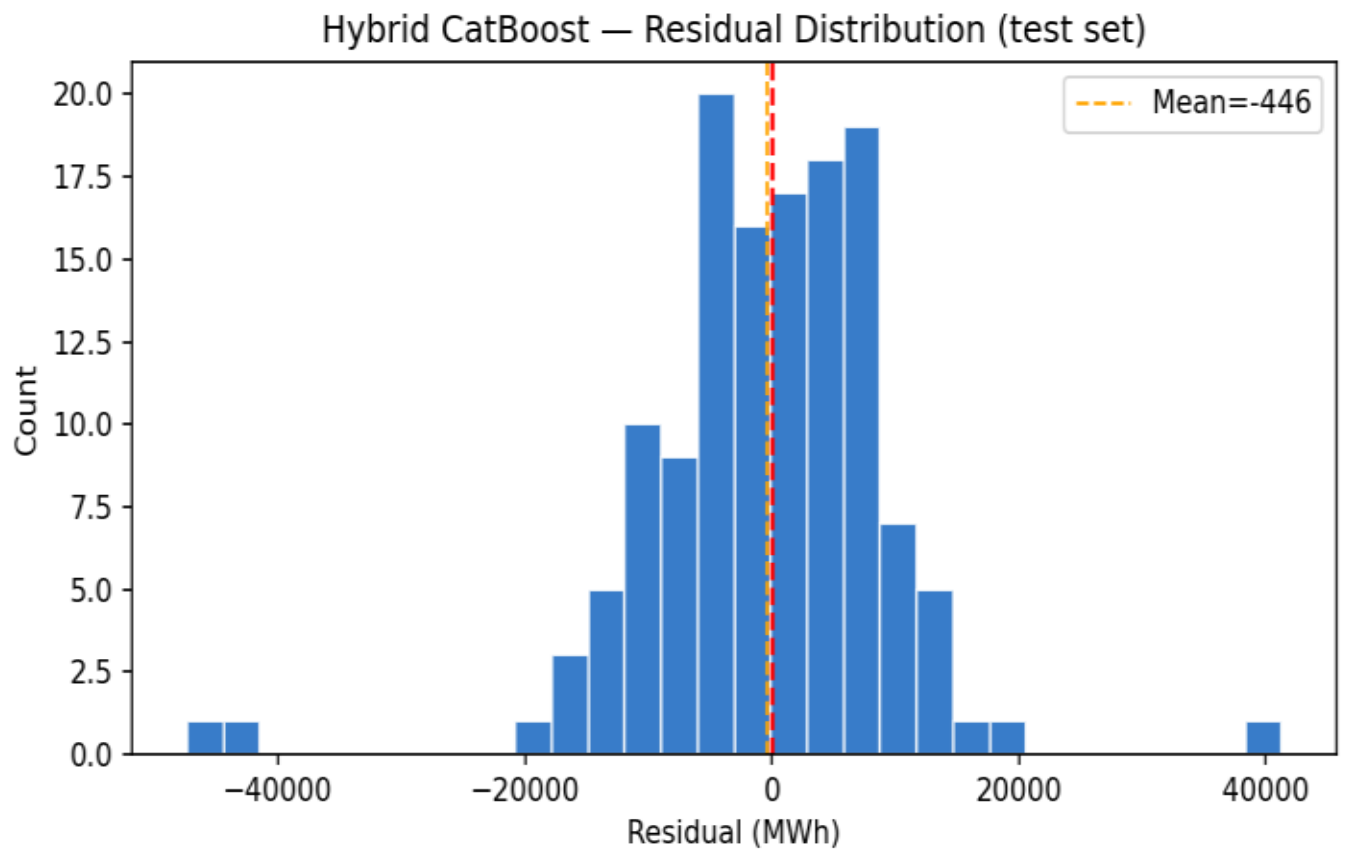

*Figure S10: Hybrid Transformer + CatBoost — Residual distribution (test set). Red dashed line: zero; orange dashed line: mean residual.*

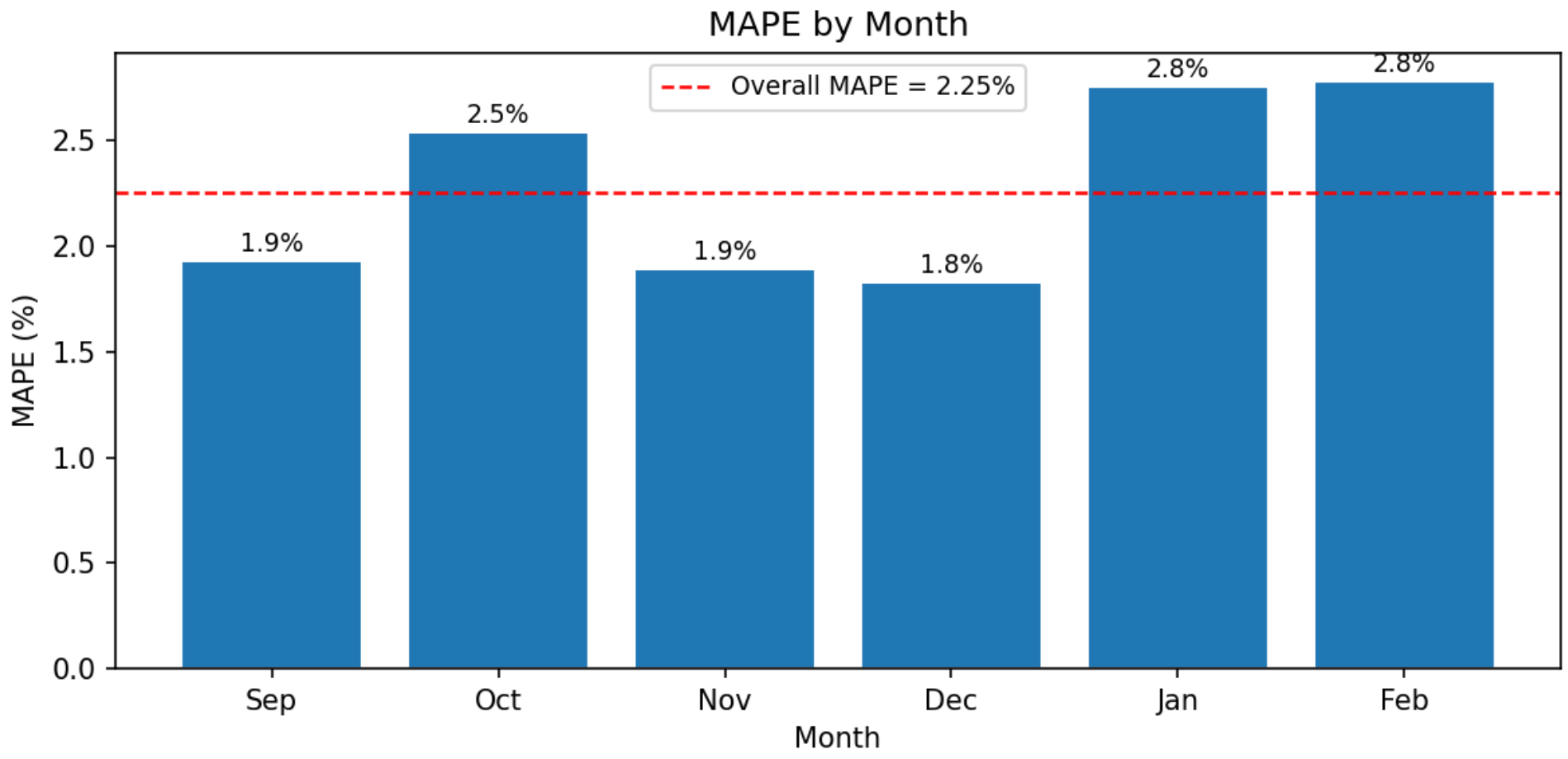

*Figure S11: Hybrid Transformer + CatBoost — Mean absolute percentage error (MAPE) by calendar month (test set). Red dashed line: overall MAPE.*

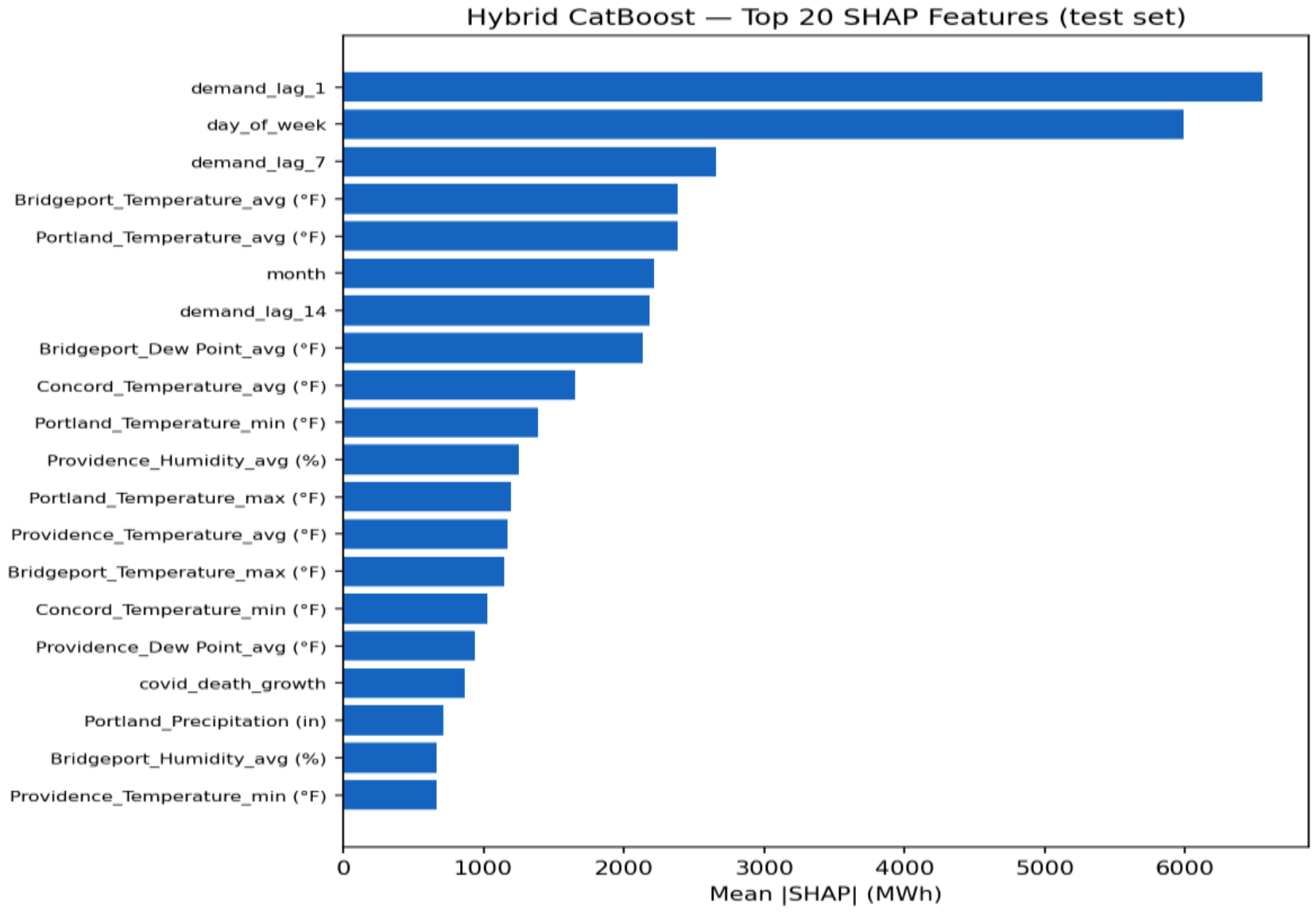

*Figure S12: Hybrid Transformer + CatBoost — Top 20 SHAP features by mean |SHAP| value (MWh) on the test set.*

## S3. Hybrid Transformer + XGBoost — Optimal Hyperparameters and Diagnostic Plots

This section presents diagnostic plots for the Hybrid Transformer + XGBoost model. The Transformer encoder (2-layer, 4-head, d_model=64, embed_dim=32) provides temporal embeddings appended to the 118 tabular features (150 total inputs).

Hybrid Transformer + XGBoost achieves an RMSE of 9,710 MWh, a MAPE of 2.22%, and an $R^2$ of 0.851 on the held-out test set (October 2022 – March 2023). All results use 300 Optuna trials (multivariate TPE sampler, seed=42) under a chronological 70/15/15 train-validation-test split.

***Table S3: Optimal hyperparameters for the Hybrid Transformer + XGBoost model (300-trial Optuna search).***

| Hyperparameter | Optimal Value | Search Range |
|---|---|---|
| n_estimators | 3,000 (early stopping) | 50 – 3,000 |
| learning_rate | 0.03892 | 0.001 – 0.3 (log) |
| max_depth | 2 | 2 – 8 |
| subsample | 0.6427 | 0.2 – 1.0 |
| colsample_bytree | 0.5416 | 0.2 – 1.0 |
| reg_alpha (L1) | $5.652 \times 10^{-6}$ | $1 \times 10^{-8}$ – 10 (log) |
| reg_lambda (L2) | 2.868 | $1 \times 10^{-8}$ – 10 (log) |

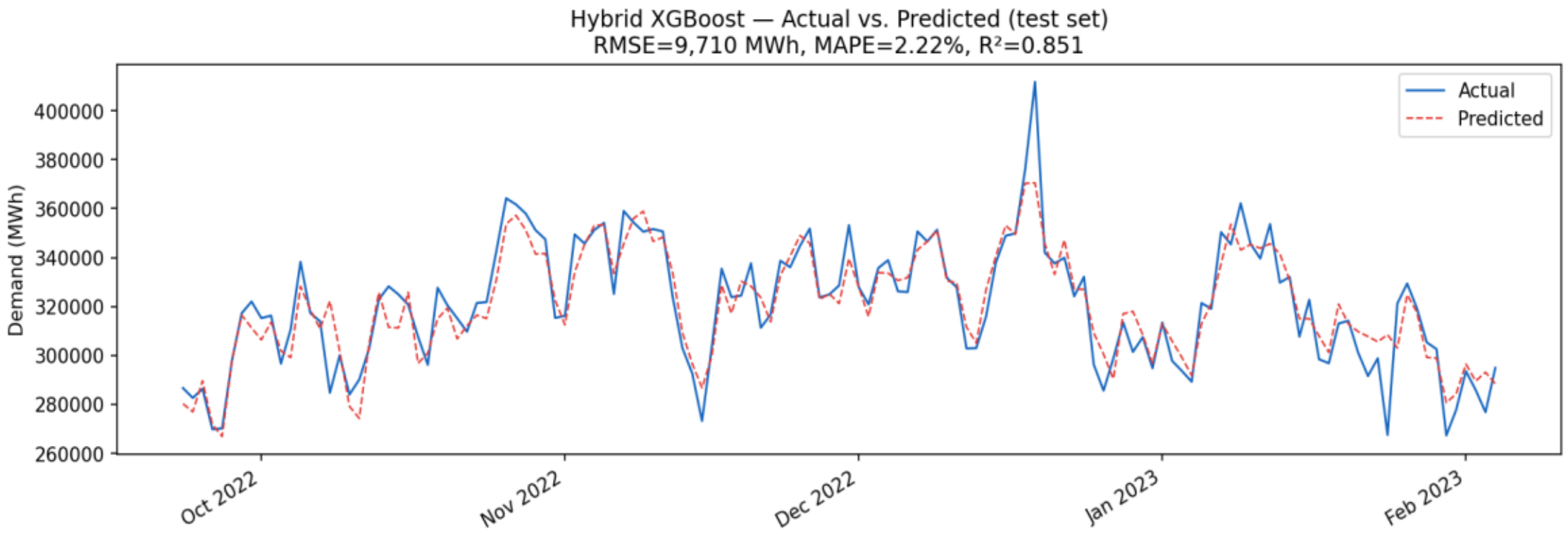

*Figure S13: Hybrid Transformer + XGBoost — Actual vs. predicted daily electricity demand (test set, October 2022 – March 2023). Blue: actual; red dashed: predicted.*

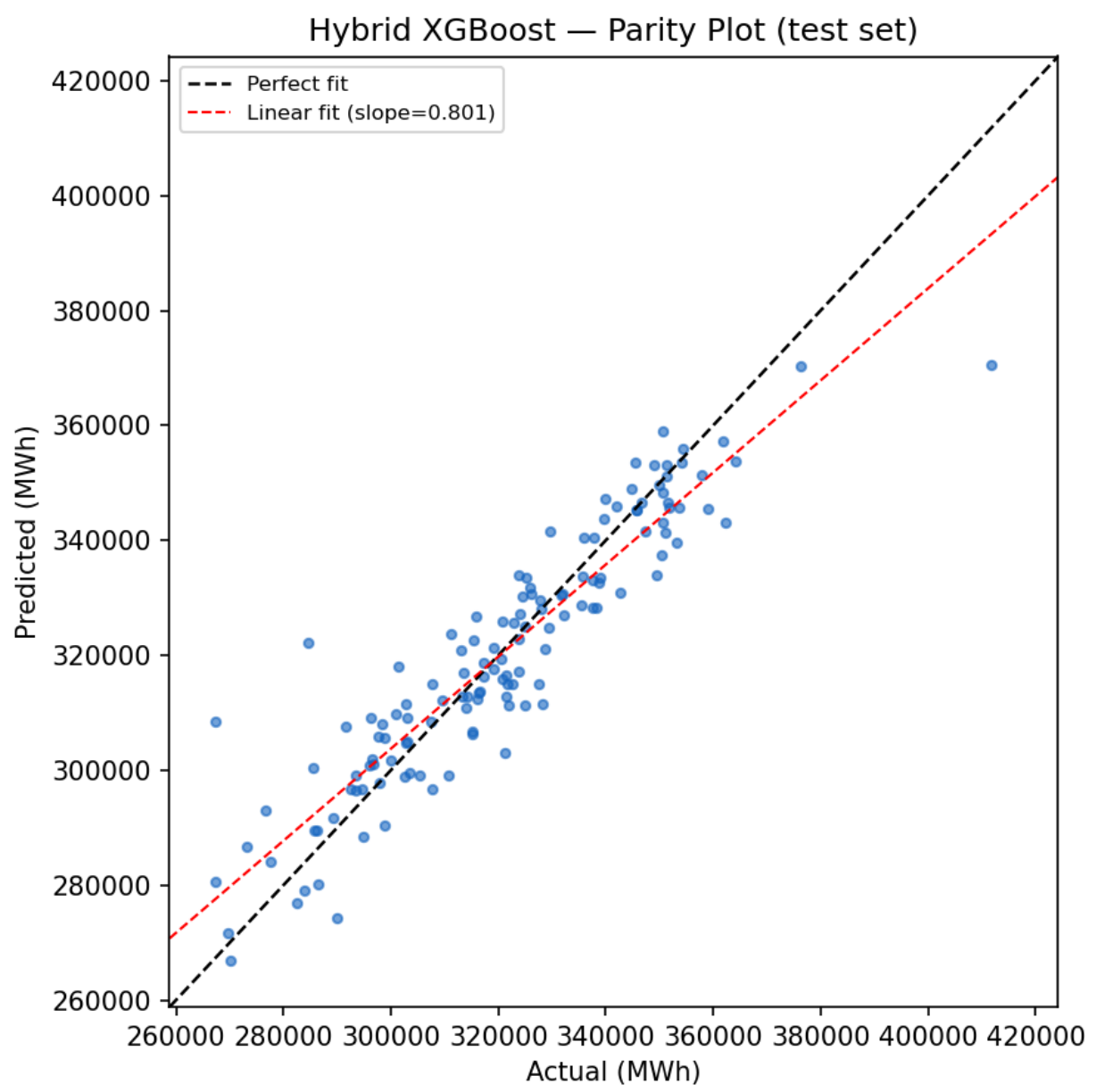

*Figure S14: Hybrid Transformer + XGBoost — Parity plot (test set). Dashed black line: perfect fit; red dashed line: linear fit to test data.*

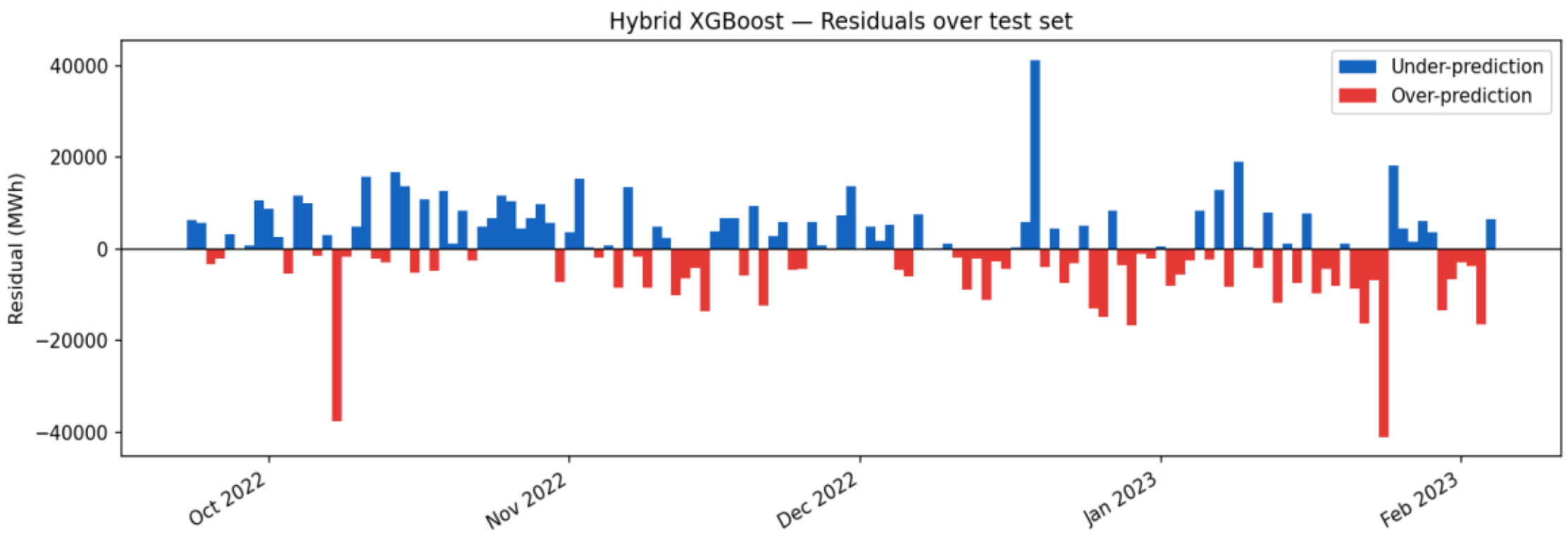

*Figure S15: Hybrid Transformer + XGBoost — Signed residuals (actual − predicted) over the test set. Blue: under-prediction; red: over-prediction.*

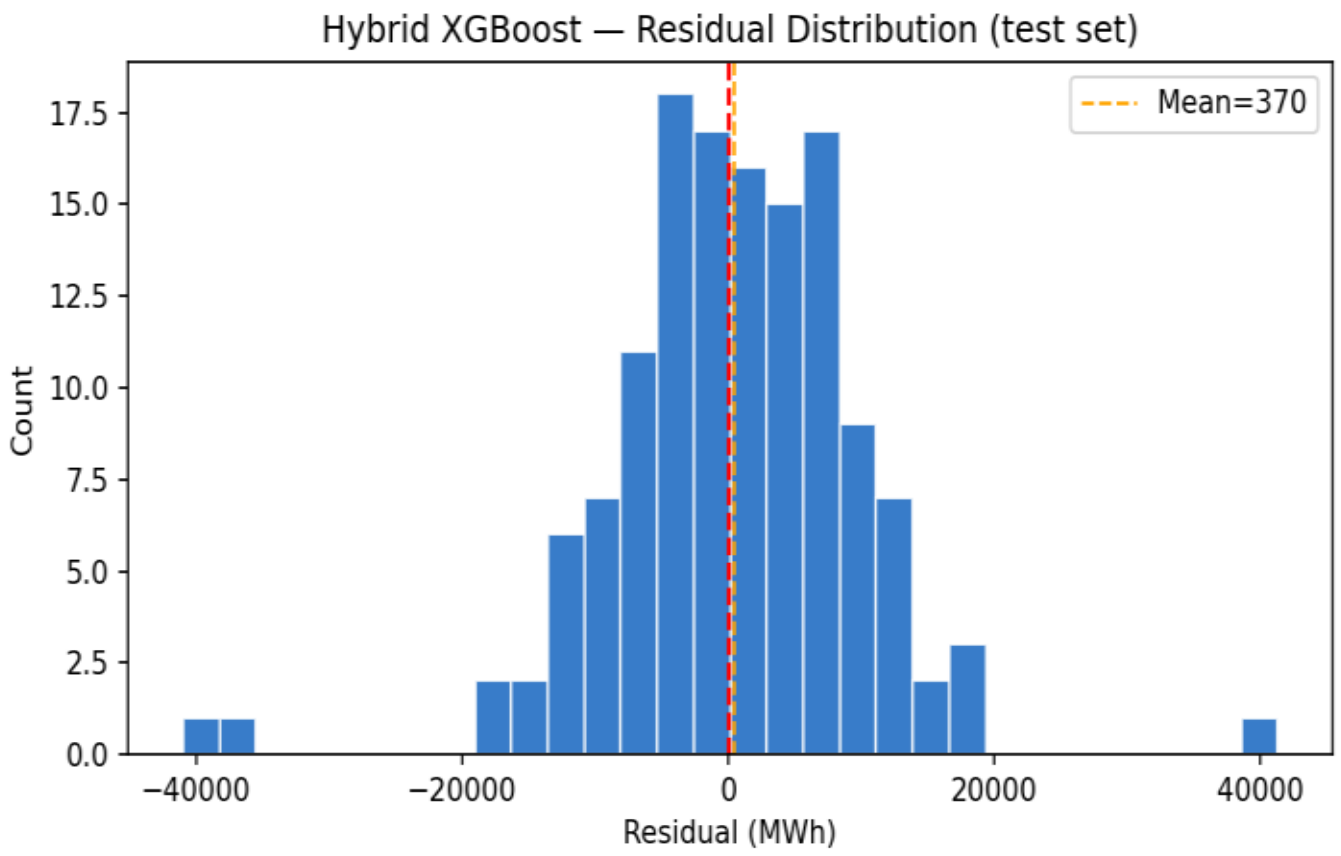

*Figure S16: Hybrid Transformer + XGBoost — Residual distribution (test set). Red dashed line: zero; orange dashed line: mean residual.*

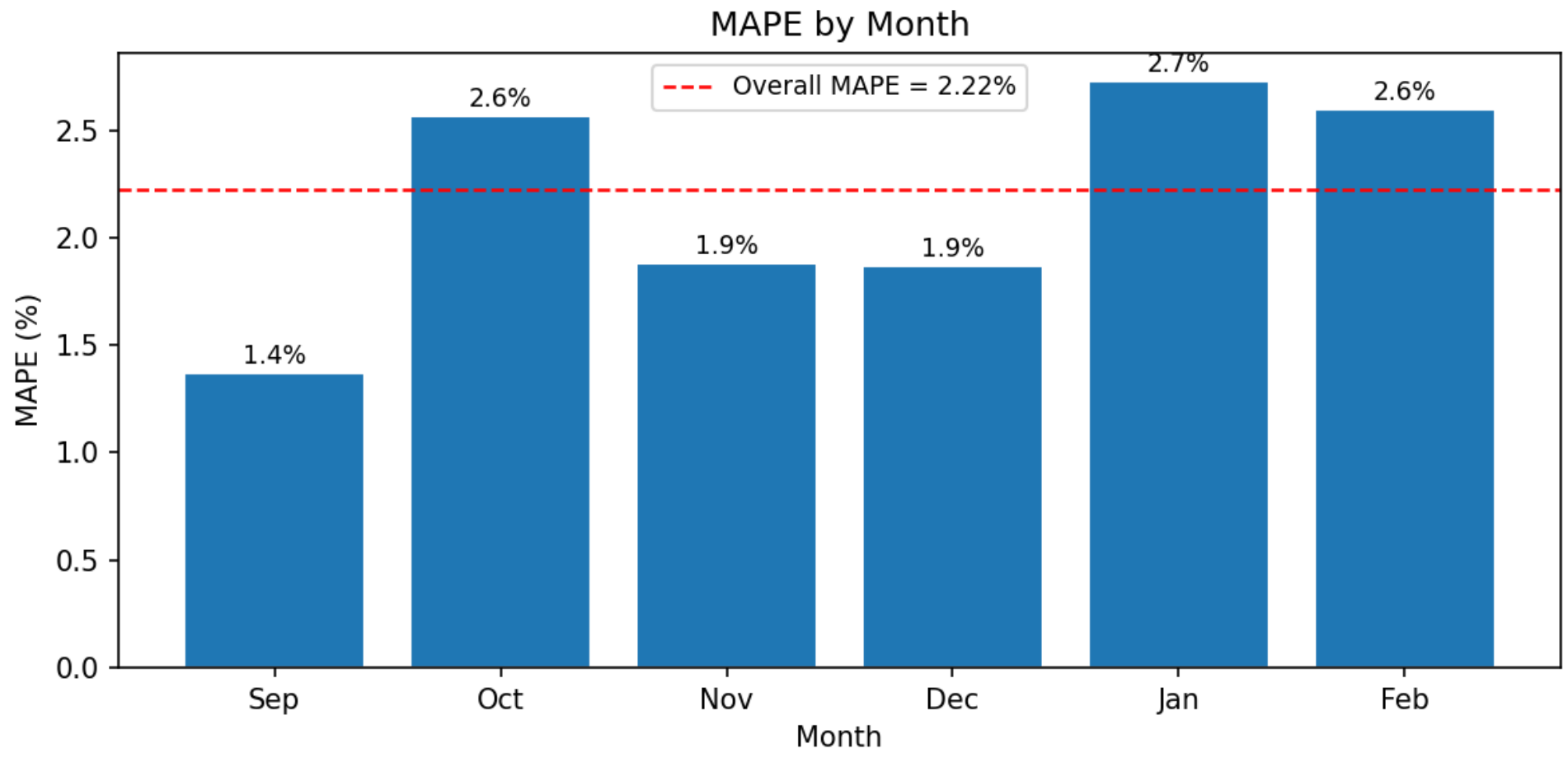

*Figure S17: Hybrid Transformer + XGBoost — Mean absolute percentage error (MAPE) by calendar month (test set). Red dashed line: overall MAPE.*

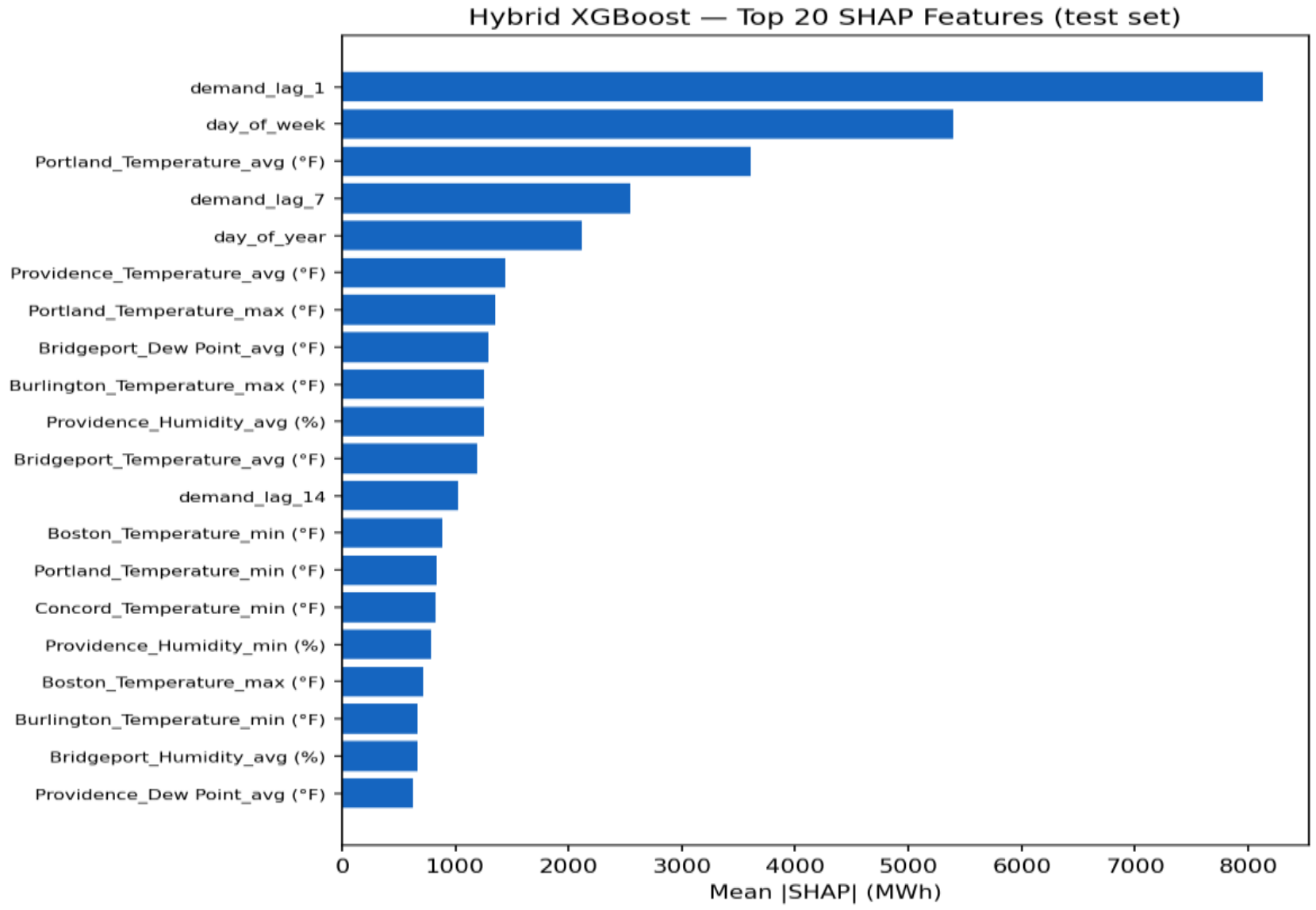

*Figure S18: Hybrid Transformer + XGBoost — Top 20 SHAP features by mean |SHAP| value (MWh) on the test set.*

## S4. Input Variable Descriptions

Table S4 summarizes all 118 input features used in the tabular models. Hybrid models use these same 118 features concatenated with 32-dimensional Transformer encoder embeddings (150 total inputs).

***Table S4: Input variable descriptions for all 118 tabular features.***

| Feature Group | Description |
|---|---|
| Weather (96 features) | Temperature (max, avg, min), dew point (max, avg, min), humidity (max, avg, min), wind speed (max, avg, min), pressure (max, avg, min), precipitation — for 6 cities: Boston, Bridgeport, Burlington, Portland, Providence, Concord. |
| COVID-19 (8 features) | Daily cases, daily deaths, 7-day rolling mean cases, 14-day rolling mean cases, case growth rate, 7-day rolling mean deaths, 14-day rolling mean deaths, death growth rate — all for New England ISO region. |
| Demand lags (4 features) | Daily electricity demand lagged by 1, 7, 14, and 30 days (MWh). |
| Calendar (6 features) | Month (1–12), day of week (0–6), day of year (1–366), season (1–4), is_summer (binary), is_winter (binary). |
| Holiday/weekend (4 features) | is_holiday, is_weekend, is_workday, is_between_holiday_weekend (binary flags from ISO-NE calendar). |